\pdfoutput=1
\documentclass[11pt,a4paper]{article}
\usepackage[]{acl}

\usepackage{times}
\usepackage{latexsym}
\usepackage{xspace}
\usepackage{graphicx}
\graphicspath{ {./images/} }
\usepackage{caption}
\usepackage{subcaption}
\usepackage{times}
\usepackage{url}
\usepackage{latexsym}
\usepackage{amsmath}
\usepackage{lipsum}
\usepackage{multirow}
\usepackage{booktabs}
\usepackage{enumerate}
\usepackage{dblfloatfix}
\usepackage{hyperref}

\usepackage{microtype}
\usepackage{float} % here for H placement parameter
\usepackage{acl}

\usepackage[T1]{fontenc}
\usepackage[utf8]{inputenc}

\title{Plumeria at SemEval-2022 Task 6: Robust Approaches for Sarcasm Detection for English and Arabic Using Transformers and Data Augmentation}

\author{Shubham Kumar Nigam$^{*}$\\
  Indian Institute of Technology Kanpur \\
  \texttt{sknigam@iitk.ac.in}\\\And
  Mosab Shaheen$^{*}$ \\
  Indian Institute of Technology Kanpur\\
  \texttt{mosab@iitk.ac.in}
 \\}

\begin{document}
\maketitle

\def\thefootnote{*}\footnotetext{These authors contributed equally to this work} \def\thefootnote{\arabic{footnote}}

\begin{abstract}
This paper describes our submission to SemEval-2022 Task 6 on sarcasm detection and its five subtasks for English and Arabic. Sarcasm conveys a meaning which contradicts the literal meaning, and it is mainly found on social networks. It has a significant role in understanding the intention of the user. For detecting sarcasm, we used deep learning techniques based on transformers due to its success in the field of Natural Language Processing (NLP) without the need for feature engineering. The datasets were taken from tweets. We created new datasets by augmenting with external data or by using word embeddings and repetition of instances. Experiments were done on the datasets with different types of preprocessing because it is crucial in this task. The rank of our team was consistent across four subtasks (fourth rank in three subtasks and sixth rank in one subtask); whereas other teams might be in the top ranks for some subtasks but rank drastically less in other subtasks. This implies the robustness and stability of the models and the techniques we used.
\end{abstract}

%Introduction
\section{Introduction}
Sarcasm is a figurative language where speakers or writers usually mean the contrary of what they say. Recognizing whether a speaker or writer is sarcastic is essential to downstream applications to understand the sentiments, opinions, and beliefs correctly \cite{ghosh-etal-2020-report}. Sarcasm is ubiquitous on the social media text and, due to its nature, can be highly divisive of computational systems that perform tasks on that kind of data such as sentiment analysis, opinion mining, and harassment detection \cite{van-hee-etal-2018-semeval, bing2012sentiment,rosenthal-etal-2014-semeval,maynard-greenwood-2014-cares}.

Our team Plumeria participated in SemEval 2022 task 6 \cite{abufarha-etal-2022-semeval} in all its subtasks on English and Arabic. Previous shared tasks on sarcasm detection \cite{VanHee2018, ghanem2019idat,ghosh-etal-2020-report,abu-farha-etal-2021-overview} have only two subtasks; one is sarcasm detection and another is predicting the type of sarcasm. However, in SemEval 2022 task 6 \cite{abufarha-etal-2022-semeval} organizers formulate three subtasks for both languages following the methods described in \cite{oprea-magdy-2020-isarcasm}.

Using the two datasets for English and Arabic, organizers formulate three subtasks as follows:
\begin{itemize}
    \item \textbf{Subtask A (English and Arabic):} It is a binary classification subtask where submitted systems have to predict whether a tweet is sarcastic or not.
    \item \textbf{Subtask B (for English only):} It is a multi-label classification subtask where submitted systems have to predict one or more labels out of six ironic-speech labels: sarcasm, irony, satire, understatement, overstatement, and rhetorical question.
    \item \textbf{Subtask C (English and Arabic):} It is a binary classification subtask. Given two texts, a sarcastic tweet and its non-sarcastic rephrase which conveys the same meaning, submitted systems have to predict which text is the sarcastic one.
\end{itemize}

    In this shared task, the systems we submitted for all three subtasks primarily focused on the transformer based approaches because of their success in the field of NLP. Moreover, we used a hierarchical network by stacking a BiLSTM layer on top of a transformer either with or without attention mechanism. However, the most significant constraint of a neural network is that they need lots of data for training to give satisfactory results. In addition, if the dataset is imbalanced with fewer instances of some class, this can result in poor results for detecting this class. That motivates us to create biased datasets, towards the concerned class/label, from existing datasets by increasing the number of instances of such a class/label. A detailed explanation of dataset creation is in section \ref{dataset}. We also illustrate the composition of created datasets in each subtask and the performance of the models on them.
    
    The rank of our team was consistent across most subtasks (fourth rank in three subtasks, sixth rank in one subtask, and tenth rank in one subtask); unlike many teams which scored high in one or two subtasks but scored considerably less in other subtasks. This shows that the robustness and the consistency of our methods. The bias in created datasets was crucial for subtask A and subtask B, while augmenting a dataset plays a key role in subtask C. We released the codes and datasets for all subtasks via GitHub\footnote{\url{https://github.com/ShubhamKumarNigam/iSarcasm-SemEval-2022-Task-6}}.

%{Related Work}
% \input{related}

%{iSarcasm  Corpus + External Corpus}
\section{Abbreviations} \label{Abbreviations}
The following abbreviations were used frequently, and are shown in Table  \ref{tab:Abbreviations}.

\begin{table}[h]
\centering
\begin{tabular}{|c|c|}
\hline
\rowcolor[HTML]{739CB9} 
{\color[HTML]{FFFFFF} \textbf{Full Form}} & {\color[HTML]{FFFFFF} \textbf{Abbreviation}} \\ \hline
\cellcolor[HTML]{FFFFFF}Language  & Lang \\ \hline
\cellcolor[HTML]{FFFFFF}Sarcastic & S    \\ \hline
Non-Sarcastic                     & NS   \\ \hline
English                           & En   \\ \hline
\cellcolor[HTML]{FFFFFF}Arabic    & Ar   \\ \hline
\cellcolor[HTML]{FFFFFF}External    & Ext   \\ \hline
\end{tabular}
\caption{Abbreviations used in the paper}
\label{tab:Abbreviations}
\end{table}

\section{Dataset} \label{dataset}

The organizers provided datasets for English and Arabic. Regardless of the dataset, these are the fields in each row of a dataset:
\begin{itemize}
    \item \textbf{Tweet}: A text specifying a tweet. This field is for all subtasks.
    \item \textbf{Sarcastic}: A binary field specifying whether a tweet is sarcastic or not. This field is for subtask A.
    \item \textbf{Rephrase}: A text specifying a non-sarcastic rephrase of a sarcastic tweet. This field is for subtask C.
    \item \textbf{Sarcasm, Irony, Satire, Understatement, Overstatement, and Rhetorical question} (English only): These binary fields are the labels of a sarcastic tweet. These fields are for subtask B.
    \item \textbf{Dialect} (Arabic only): A text specifying the dialect of a tweet from one of five dialects: Modern Standard Arabic (MSA), Egyptian, Levantine, Maghrebi, and Gulf. This field is for subtask A and C.
\end{itemize}

In the following sections we will describe the datasets given by the organizers, other available datasets, and augmented datasets.

\subsection{Datasets given by organisers (Original)}
The organizers released two training datasets for Arabic and English to train the systems on them for all subtasks. Later on, they released one testing dataset for each subtask. We called these datasets "original" datasets as they are the official datasets for the subtasks. Information about the distribution of sarcastic and non-sarcastic tweets is presented in Table \ref{tab:TaskA-Datasets} and in the appendix \ref{More_Information} - Figure \ref{fig:OriginalEnglishArabicTaskAC}. Furthermore, information about sarcastic labels for subtask B is presented in Table \ref{tab:TaskB-Datasets} and in the appendix \ref{More_Information} - Figure \ref{fig:OriginalEnglishTaskB}.

\subsection{Other Available Datasets (External)}
The datasets in this section are not the official datasets and thus we called them "external" datasets. However, we used these datasets for subtask A and subtask B as they are created for similar subtasks.

\textbf{Datasets Downloaded using Twitter API:}
Initially the organizers provided the participants with a training and testing datasets which covered subtask A and subtask B for English. However, they provided the tweet ID instead of the tweet text and they asked the participants to download the tweet text using the tweet ID via Twitter API\footnote{\href{https://developer.twitter.com/en/docs/api-reference-index}{developer.twitter.com/en/docs/api-reference-index}}. Therefore, we downloaded some of the tweet texts for these two datasets and other tweet texts we could not find. We were able to download 2841 tweets for training and 713 tweets for testing as shown in Table \ref{tab:TaskA-Datasets}. The distribution of the tweets over the sarcastic labels is presented in Table \ref{tab:TaskB-Datasets}.

\textbf{Datasets of SemEval-2018 Task 3:}
These datasets are on same subtasks of subtask A and subtask B but for SemEval 2018 \citep{VanHee2018}. We used the dataset for subtask A which has emojis and sarcasm hashtags. The datasets are available for download in this link\footnote{\href{https://github.com/Cyvhee/SemEval2018-Task3/tree/master/datasets/train}{SemEval2018-Task3 Dataset}}. More information about the distribution of sarcastic and non-sarcastic tweets is presented in Table \ref{tab:TaskA-Datasets}.

\textbf{ArSarcasm-v2 Dataset:}
It contains a training and testing datasets for sarcasm detection in Arabic \citep{abu-farha-etal-2021-overview}. Each row contains a tweet, sarcastic class, sentiment, and dialect. The datasets are available for download in this link\footnote{\href{https://github.com/iabufarha/ArSarcasm-v2}{ArSarcasm-v2}}. More information about the distribution of sarcastic and non-sarcastic tweets is presented in Table \ref{tab:TaskA-Datasets}.

%%%%%%%%%%% Task - A Datasets %%%%%%%%%%%%%% 

\begin{table}[h]
\centering
\resizebox{\columnwidth}{!}{
\begin{tabular}{|l|
>{\columncolor[HTML]{FFFFFF}}l |l|l|l|l|}
\hline
\cellcolor[HTML]{739CB9}{\color[HTML]{FFFFFF} \textbf{Dataset}} &
  \cellcolor[HTML]{739CB9}{\color[HTML]{FFFFFF} \textbf{Split}} &
  \cellcolor[HTML]{739CB9}{\color[HTML]{FFFFFF} \textbf{Lang}} &
  \cellcolor[HTML]{739CB9}{\color[HTML]{FFFFFF} \textbf{Total}} &
  \cellcolor[HTML]{739CB9}{\color[HTML]{FFFFFF} \textbf{S\%}} &
  \cellcolor[HTML]{739CB9}{\color[HTML]{FFFFFF} \textbf{NS\%}} \\ \hline
Original     & Train & En                         & 3468  & 25   & 75   \\ \hline
Original     & Train & Ar                         & 3102  & 24   & 76   \\ \hline
Original     & Test  & En                         & 1400  & 14.3 & 85.7 \\ \hline
Original     & Test  & Ar                         & 1400  & 14.3 & 85.7 \\ \hline
Twitter API  & Train & En                         & 2841  & 16.8 & 83.2 \\ \hline
Twitter API  & Test  & En                         & 713   & 16.8 & 83.2 \\ \hline
SemEval 2018 & Train & \cellcolor[HTML]{FFFFFF}En & 3834  & 49.8 & 50.2 \\ \hline
ArSarcasm-v2 & Train & \cellcolor[HTML]{FFFFFF}Ar & 12548 & 17.3 & 82.7 \\ \hline
ArSarcasm-v2 & Test  & Ar                         & 3000  & 27.4 & 72.6 \\ \hline
\end{tabular}
}
\caption{Total number of tweets and percentage of sarcastic and non-sarcastic tweets in each dataset for subtask A.}
\label{tab:TaskA-Datasets}
\end{table}

%%%%%%%%%%% Task - B Datasets %%%%%%%%%%%%%%

\begin{table}[h]
\resizebox{\columnwidth}{!}{
\begin{tabular}{|c|c|c|c|c|c|}
\hline
\rowcolor[HTML]{739CB9} 
\cellcolor[HTML]{739CB9}{\color[HTML]{FFFFFF} } &
  \cellcolor[HTML]{739CB9}{\color[HTML]{FFFFFF} } &
  \cellcolor[HTML]{739CB9}{\color[HTML]{FFFFFF} } &
  {\color[HTML]{FFFFFF} \textbf{Sarcasm\%}} &
  {\color[HTML]{FFFFFF} \textbf{Irony\%}} &
  {\color[HTML]{FFFFFF} \textbf{Satire\%}} \\ \cline{4-6} 
\rowcolor[HTML]{739CB9} 
\multirow{-2}{*}{\cellcolor[HTML]{739CB9}{\color[HTML]{FFFFFF} \textbf{Dataset}}} &
  \multirow{-2}{*}{\cellcolor[HTML]{739CB9}{\color[HTML]{FFFFFF} \textbf{Split}}} &
  \multirow{-2}{*}{\cellcolor[HTML]{739CB9}{\color[HTML]{FFFFFF} \textbf{Total}}} &
  {\color[HTML]{FFFFFF} \textbf{\begin{tabular}[c]{@{}c@{}}Under-\\ statement\%\end{tabular}}} &
  {\color[HTML]{FFFFFF} \textbf{\begin{tabular}[c]{@{}c@{}}Over-\\ statement\%\end{tabular}}} &
  {\color[HTML]{FFFFFF} \textbf{\begin{tabular}[c]{@{}c@{}}Rhetorical\\ question\%\end{tabular}}} \\ \hline
                                                                         & \cellcolor[HTML]{FFFFFF}                        &                        & 68.3 & 14.8 & 2.4  \\ \cline{4-6} 
                                                                         & \multirow{-2}{*}{\cellcolor[HTML]{FFFFFF}Train} & \multirow{-2}{*}{3468} & 1    & 3.8  & 9.7  \\ \cline{2-6} 
                                                                         & \cellcolor[HTML]{FFFFFF}                        &                        & 66.4 & 7.4  & 18.1 \\ \cline{4-6} 
\multirow{-4}{*}{Original}                                               & \multirow{-2}{*}{\cellcolor[HTML]{FFFFFF}Test}  & \multirow{-2}{*}{1400} & 0.4  & 3.7  & 4    \\ \hline
                                                                         & \cellcolor[HTML]{FFFFFF}                        &                        & 41.5 & 30.6 & 10.9 \\ \cline{4-6} 
                                                                         & \multirow{-2}{*}{\cellcolor[HTML]{FFFFFF}Train} & \multirow{-2}{*}{2841} & 1    & 8.2  & 7.8  \\ \cline{2-6} 
                                                                         & \cellcolor[HTML]{FFFFFF}                        &                        & 41.7 & 33.3 & 11.7 \\ \cline{4-6} 
\multirow{-4}{*}{\begin{tabular}[c]{@{}c@{}}Twitter \\ API\end{tabular}} & \multirow{-2}{*}{\cellcolor[HTML]{FFFFFF}Test}  & \multirow{-2}{*}{713}  & 3.3  & 6.7  & 3.3  \\ \hline
\end{tabular}
}
\caption{Total number of tweets and percentage of sarcastic labels in each dataset for subtask B.}
\label{tab:TaskB-Datasets}
\end{table}

\subsection{Augmented Datasets}
In addition to the original and external datasets, we created more datasets with more number of instances using the following methods:

\begin{enumerate}
  \item \textbf{Augmenting an original dataset with external datasets:}
    We added instances to an original dataset from the matching external datasets either to balance it or just to augment it, and we filtered out the NAN entries .
  \item \textbf{Augmenting a dataset using word embeddings:}
    For word embeddings we used Gensim library\footnote{\href{https://radimrehurek.com/gensim/}{Gensim Library}} together with GloVe word vectors\footnote{\href{https://github.com/RaRe-Technologies/gensim-data}{GloVe Word Vectors}} trained on two billion tweets with 100 dimension word vectors (glove-twitter-100). To create new instances in a dataset, we took a copy of one instance in the dataset and replaced up to four keywords in a tweet (or its rephrase) with one of the top three similar words according to the similarity between the word vectors, then we added the copied modified instance to the dataset and we repeated the process for other instances.
  \item \textbf{Augmenting a dataset by repeating instances:}
    Instances from a dataset are mostly repeated to balance the classes/labels in a dataset.
\end{enumerate}

The final datasets used for each subtask is explained in the dedicated section for the subtask.

%{iSarcasm  Corpus + External Corpus}

\section{Preprocessing} \label{preprocessing}
%Four versions of the final datasets were created using four types of preprocessing as follows:%

\begin{itemize}
  \item \textbf{Type I:} No preprocessing
  \item \textbf{Type II:} 
  Emotion icons were converted to their string text using the "emoji" Python library. Then, URLs were converted to  "HTTPURL" token, also every mention in a tweet was converted to "@USER" token using regular expressions. These conversions was done because the BERTweet model was pre-trained on tweets after these conversions.
  \item \textbf{Type III:}
  Same as in Type-III besides converting the smiley face codes e.g. ":‑)" and ":)" to one of three values (smiley, sad, and playful). More than two successive occurrences of any punctuation like in "why?!!!!" were removed, then we removed more than two successive occurrences of same character like in "Superrrr" which can be found frequently in tweets. Moreover, a contraction (e.g. "isn't and "'cause") was replaced with its full form (e.g. "is not" and "because").
  \item \textbf{Type IV:}
  Same as in Type-III besides stemming and stop-word removal. For English we used WordNet lemmatizer and for Arabic we used ISRI stemmer, all from the NLTK Python library\footnote{\href{https://www.nltk.org/}{NLTK Python Library}}.
\end{itemize}

%Experiments
\section{Methodology, Results and Analysis}

\subsection{Conventions} \label{Conventions}
In the following tables, if a table cell is highlighted with a light brown color, it means that the score is among the best results on the validation dataset in the corresponding section of the table; whereas the one with brown color is the highest score. Furthermore, if a cell is highlighted with a green color, it means that the score is our final submission score (released by the organizers) in the subtask on the test dataset; whereas the one with blue color is a score of a submitted model but not the final one (a team can have multiple submissions).

%%%%%%%%%%%%%%%%%%%%%%%%%%%%%%%%%%%%%%%%%%%%%%
%%%%%%%%%%%%%%%% subtask A English %%%%%%%%%%%%%%
%%%%%%%%%%%%%%%%%%%%%%%%%%%%%%%%%%%%%%%%%%%%%%

\subsection{Subtask A (English)}

\subsubsection{Datasets}
\begin{itemize}
  \item \textbf{Original}: The training split of the original dataset for English in Table \ref{tab:TaskA-Datasets} is splitted into training and validation datasets as shown in Table \ref{tab:TaskA-En-Original-Datasets}.
  \item \textbf{External}: As the measure for this task is F1-score for the sarcastic class, thus we created datasets which are biased towards the sarcastic class as shown in Table
\ref{tab:TaskA-En-External-Biased-Datasets}.
\end{itemize}

%%%%%%%%%%%%% TaskA-En-Original-Datasets %%%%%%%%%%

\begin{table}[h]
\centering
\resizebox{0.8\columnwidth}{!}{
\begin{tabular}{|c|
>{\columncolor[HTML]{FFFFFF}}c |
>{\columncolor[HTML]{FFFFFF}}c |
>{\columncolor[HTML]{FFFFFF}}c |}
\hline
\cellcolor[HTML]{739CB9}{\color[HTML]{FFFFFF} \textbf{Dataset}} &
  \cellcolor[HTML]{739CB9}{\color[HTML]{FFFFFF} \textbf{Total}} &
  \cellcolor[HTML]{739CB9}{\color[HTML]{FFFFFF} \textbf{S\%}} &
  \cellcolor[HTML]{739CB9}{\color[HTML]{FFFFFF} \textbf{NS\%}} \\ \hline
Original Train &
  2080 &
  25 &
  75 \\ \hline
Original Val &
  1388 &
  25 &
  75 \\ \hline
\end{tabular}
}
\caption{Original datasets for subtask A (English) with the total number of tweets and the percentage of sarcastic (S\%) and non-sarcastic (NS\%) tweets.}
\label{tab:TaskA-En-Original-Datasets}
\end{table}

%%%%%%%%%%%% Biased external datasets for subtask A (English) %%%%%%%%%%%%
%%%%%%%%%%%%%%%%%%%%%%%%%%%%%%%%%%%%%%%%%%%%%%%%%%%%%%%%%%%%%%%%%%%%%%%

\begin{table}[h]
\resizebox{\columnwidth}{!}{
\begin{tabular}{|
>{\columncolor[HTML]{FFFFFF}}c |c|
>{\columncolor[HTML]{FFFFFF}}c |
>{\columncolor[HTML]{FFFFFF}}c |c|c|}
\hline
\cellcolor[HTML]{739CB9}{\color[HTML]{FFFFFF} \textbf{Dataset}} &
  \cellcolor[HTML]{739CB9}{\color[HTML]{FFFFFF} \textbf{\begin{tabular}[c]{@{}c@{}}Contributing\\ Datasets\end{tabular}}} &
  \cellcolor[HTML]{739CB9}{\color[HTML]{FFFFFF} \textbf{\begin{tabular}[c]{@{}c@{}}Additional Non\\ Sarcastic Tweets\\ from SemEval\\ 2018-Training\end{tabular}}} &
  \cellcolor[HTML]{739CB9}{\color[HTML]{FFFFFF} \textbf{Total}} &
  \cellcolor[HTML]{739CB9}{\color[HTML]{FFFFFF} \textbf{S\%}} &
  \cellcolor[HTML]{739CB9}{\color[HTML]{FFFFFF} \textbf{NS\%}} \\ \hline
B0 &  & 0    & 4578 & \cellcolor[HTML]{FFFFFF}66 & 34                         \\ \cline{1-1} \cline{3-6} 
B1 &  & 145  & 4723 & 64                         & 36                         \\ \cline{1-1} \cline{3-6} 
B2 &  & 290  & 4868 & \cellcolor[HTML]{FFFFFF}62 & \cellcolor[HTML]{FFFFFF}38 \\ \cline{1-1} \cline{3-6} 
B3 &  & 435  & 5013 & \cellcolor[HTML]{FFFFFF}60 & \cellcolor[HTML]{FFFFFF}40 \\ \cline{1-1} \cline{3-6} 
B4 &  & 580  & 5158 & \cellcolor[HTML]{FFFFFF}59 & \cellcolor[HTML]{FFFFFF}41 \\ \cline{1-1} \cline{3-6} 
B5 &  & 725  & 5303 & \cellcolor[HTML]{FFFFFF}57 & \cellcolor[HTML]{FFFFFF}43 \\ \cline{1-1} \cline{3-6} 
B6 &  & 870  & 5448 & \cellcolor[HTML]{FFFFFF}55 & \cellcolor[HTML]{FFFFFF}45 \\ \cline{1-1} \cline{3-6} 
B7 &  & 1015 & 5593 & \cellcolor[HTML]{FFFFFF}54 & \cellcolor[HTML]{FFFFFF}46 \\ \cline{1-1} \cline{3-6} 
B8 &  & 1160 & 5738 & \cellcolor[HTML]{FFFFFF}53 & \cellcolor[HTML]{FFFFFF}47 \\ \cline{1-1} \cline{3-6} 
B9 &
  \multirow{-10}{*}{\begin{tabular}[c]{@{}c@{}}Original Train +\\ Twitter API Train\\ (only sarcastic) +\\ Twitter API Test \\ (only sarcastic) +\\ SemEval 2018 Train\\ (1911 sarcastic)\end{tabular}} &
  1305 &
  5883 &
  \cellcolor[HTML]{FFFFFF}51 &
  \cellcolor[HTML]{FFFFFF}49 \\ \hline
\end{tabular}
}
\caption{External datasets for subtask A (English).}
\label{tab:TaskA-En-External-Biased-Datasets}
\end{table}

%%%%%%%%%%%%%%%%%%%%%%%%%%%%%%%%%%%%%%%%%%%%%%%%%%%%%%%%%%%%%%%%%%%%%%%

\subsubsection{Approaches}
We primarily focused on the transformer based models. Since the task is a binary classification on tweets, the excellent choice to start with is BERTweet-base\footnote{\href{https://huggingface.co/vinai/bertweet-base}{HuggingFace Bertweet-Base}} and BERTweet-large\footnote{\href{https://huggingface.co/vinai/bertweet-large}{HuggingFace Bertweet-Large}} \cite{bertweet}, a pre-trained language model on 845M English Tweets. Likewise, we tried the ELECTRA\footnote{\href{https://huggingface.co/google/electra-large-discriminator}{HuggingFace Electra Large Discriminator}} \cite{clark2020electra} replaced token detection model (a pre-training task in which the model learns to distinguish real input tokens). In ELECTRA model, some tokens in the input are replaced with sample tokens instead of masking the tokens as in BERT. Moreover, we used a hierarchical network by passing the input tokens to the BERT model, then each token embedding is passed to a Bi-LSTM layer either with or without attention. The architecture of the BERT model, ELECTRA model, and hierarchical network is shown in appendix \ref{BERT-arch}, \ref{ELECTRA-arch}, and \ref{Hier_Network} respectively. The final layer of each model was a linear layer with softmax activation function and we used the cross entropy loss function.

\textbf{Note:} We ran several experiments on all the approaches of this subtask with different datasets and preprocessing types. We also experimented with different learning rates, epochs, and loss functions to verify which one is performing best. 

%%%%%%%%%%%%%%%%%%%%%%%%%%%%%%%%%%%

\subsubsection{Results and Analysis}
Metric: The main metric is F1-score for the sarcastic class.

\textbf{BERT:}
We used BERTweet-large, as it gave better performance than BERTweet-base, on the original training dataset shown in Table \ref{tab:TaskA-En-Original-Datasets}. We experimented with different learning rates and preprocessing types, and ran for 5 epochs. The results are shown in Table \ref{tab:TaskA-En-Original-Results}.

\begin{table}[h]
\centering

\begin{tabular}{|
>{\columncolor[HTML]{FFFFFF}}c |c|
>{\columncolor[HTML]{FFFFFF}}c |
>{\columncolor[HTML]{FFFFFF}}c |}
\hline
\cellcolor[HTML]{739CB9}{\color[HTML]{FFFFFF} \textbf{\begin{tabular}[c]{@{}c@{}}Learning\\ Rate\end{tabular}}} &
  \cellcolor[HTML]{739CB9}{\color[HTML]{FFFFFF} \textbf{Type}} &
  \cellcolor[HTML]{739CB9}{\color[HTML]{FFFFFF} \textbf{Val}} &
  \cellcolor[HTML]{739CB9}{\color[HTML]{FFFFFF} \textbf{Test}} \\ \hline
\cellcolor[HTML]{FFFFFF}                           & I                          & 0.0057                         & 0.0293 \\ \cline{2-4} 
\cellcolor[HTML]{FFFFFF}                           & \cellcolor[HTML]{FFFFFF}II & \cellcolor[HTML]{EADDCA}0.5017 & 0.457  \\ \cline{2-4} 
\cellcolor[HTML]{FFFFFF}                           & III                        & 0                              & 0      \\ \cline{2-4} 
\multirow{-4}{*}{\cellcolor[HTML]{FFFFFF}2 e - 6} & IV                         & 0                              & 0      \\ \hline
\cellcolor[HTML]{FFFFFF}                           & I                          & 0.3786                         & 0.5068 \\ \cline{2-4} 
\cellcolor[HTML]{FFFFFF}                           & \cellcolor[HTML]{FFFFFF}II & \cellcolor[HTML]{EADDCA}0.5552 & 0.4874 \\ \cline{2-4} 
\cellcolor[HTML]{FFFFFF}                           & III                        & \cellcolor[HTML]{EADDCA}0.5405 & 0.4972 \\ \cline{2-4} 
\multirow{-4}{*}{\cellcolor[HTML]{FFFFFF}3 e - 6} & IV                         & 0                              & 0      \\ \hline
\cellcolor[HTML]{FFFFFF}                           & I                          & \cellcolor[HTML]{EADDCA}0.5585 & 0.4981 \\ \cline{2-4} 
\cellcolor[HTML]{FFFFFF}                           & \cellcolor[HTML]{FFFFFF}II & 0.4926                         & 0.4717 \\ \cline{2-4} 
\cellcolor[HTML]{FFFFFF}                           & III                        & \cellcolor[HTML]{EADDCA}0.5407 & 0.4772 \\ \cline{2-4} 
\multirow{-4}{*}{\cellcolor[HTML]{FFFFFF}4 e - 6} & IV                         & 0                              & 0      \\ \hline
\cellcolor[HTML]{FFFFFF}                           & I                          & \cellcolor[HTML]{EADDCA}0.5275 & 0.4724 \\ \cline{2-4} 
\cellcolor[HTML]{FFFFFF}                           & \cellcolor[HTML]{FFFFFF}II & \cellcolor[HTML]{EADDCA}0.5655 & 0.4841 \\ \cline{2-4} 
\cellcolor[HTML]{FFFFFF}                           & III                        & 0                              & 0      \\ \cline{2-4} 
\multirow{-4}{*}{\cellcolor[HTML]{FFFFFF}5 e - 6} & IV                         & 0                              & 0      \\ \hline
\end{tabular}

\caption{Results of the BERT model for subtask A (English) on original datasets}
\label{tab:TaskA-En-Original-Results}
\end{table}

From these experiments we found that Type-II preprocessing is performing better than other types and same applies for the learning rate 4e-6. We conducted similar experiments on the external datasets in Table \ref{tab:TaskA-En-External-Biased-Datasets} and we found similar results. We tried using the cross entropy loss function with and without weights on the external datasets using the same learning rate, preprocessing type, and number of epochs. We got our best result on the B4 dataset with weighted loss function which was our official submission score for this subtask. The results are shown in Table \ref{tab:TaskA-En-Biased-Results}.

\begin{table}[h]
\centering
\resizebox{\columnwidth}{!}{
\begin{tabular}{|c|
>{\columncolor[HTML]{FFFFFF}}c 
>{\columncolor[HTML]{FFFFFF}}c |
>{\columncolor[HTML]{FFFFFF}}c 
>{\columncolor[HTML]{FFFFFF}}c |}
\hline
\cellcolor[HTML]{739CB9}{\color[HTML]{FFFFFF} } &
  \multicolumn{2}{c|}{\cellcolor[HTML]{739CB9}{\color[HTML]{FFFFFF} \textbf{\begin{tabular}[c]{@{}c@{}}Loss1: Without\\ Weights\end{tabular}}}} &
  \multicolumn{2}{c|}{\cellcolor[HTML]{739CB9}{\color[HTML]{FFFFFF} \textbf{\begin{tabular}[c]{@{}c@{}}Loss2: W1=1/\#NS,\\W2=1/\#S\end{tabular}}}} \\ \cline{2-5} 
\multirow{-3}{*}{\cellcolor[HTML]{739CB9}{\color[HTML]{FFFFFF} \textbf{Biased}}} &
  \multicolumn{1}{c|}{\cellcolor[HTML]{739CB9}{\color[HTML]{FFFFFF} \textbf{Val}}} &
  \cellcolor[HTML]{739CB9}{\color[HTML]{FFFFFF} \textbf{Test}} &
  \multicolumn{1}{c|}{\cellcolor[HTML]{739CB9}{\color[HTML]{FFFFFF} \textbf{Val}}} &
  \cellcolor[HTML]{739CB9}{\color[HTML]{FFFFFF} \textbf{Test}} \\ \hline
B0 &
  \multicolumn{1}{c|}{\cellcolor[HTML]{FFFFFF}0.5784} &
  0.4487 &
  \multicolumn{1}{c|}{\cellcolor[HTML]{FFFFFF}0.5944} &
  0.4519 \\ \hline
B1 &
  \multicolumn{1}{c|}{\cellcolor[HTML]{FFFFFF}0.5714} &
  0.4548 &
  \multicolumn{1}{c|}{\cellcolor[HTML]{FFFFFF}0.5738} &
  0.479 \\ \hline
B2 &
  \multicolumn{1}{c|}{\cellcolor[HTML]{FFFFFF}0.5767} &
  0.465 &
  \multicolumn{1}{c|}{\cellcolor[HTML]{FFFFFF}0.5951} &
  0.4917 \\ \hline
B3 &
  \multicolumn{1}{c|}{\cellcolor[HTML]{D5AB6D}0.601} &
  \cellcolor[HTML]{6D9EEB}0.4626 &
  \multicolumn{1}{c|}{\cellcolor[HTML]{FFFFFF}0.5931} &
  0.4817 \\ \hline
B4 &
  \multicolumn{1}{c|}{\cellcolor[HTML]{FFFFFF}0.5954} &
  0.4727 &
  \multicolumn{1}{c|}{\cellcolor[HTML]{D5AB6D}0.6025} &
  \cellcolor[HTML]{6AA84F}0.4769 \\ \hline
B5 &
  \multicolumn{1}{c|}{\cellcolor[HTML]{FFFFFF}0.5874} &
  0.5142 &
  \multicolumn{1}{c|}{\cellcolor[HTML]{FFFFFF}0.5803} &
  0.5008 \\ \hline
B6 &
  \multicolumn{1}{c|}{\cellcolor[HTML]{FFFFFF}0.5858} &
  0.4791 &
  \multicolumn{1}{c|}{\cellcolor[HTML]{FFFFFF}0.5624} &
  0.4884 \\ \hline
B7 &
  \multicolumn{1}{c|}{\cellcolor[HTML]{FFFFFF}0.5957} &
  0.5016 &
  \multicolumn{1}{c|}{\cellcolor[HTML]{FFFFFF}0.5637} &
  0.5034 \\ \hline
B8 &
  \multicolumn{1}{c|}{\cellcolor[HTML]{FFFFFF}0.584} &
  0.492 &
  \multicolumn{1}{c|}{\cellcolor[HTML]{FFFFFF}0.5814} &
  0.5 \\ \hline
B9 &
  \multicolumn{1}{c|}{\cellcolor[HTML]{FFFFFF}0.5723} &
  0.487 &
  \multicolumn{1}{c|}{\cellcolor[HTML]{FFFFFF}0.5554} &
  0.49  \\ \hline
\end{tabular}
}
\caption{Results of the BERT model for subtask A (English) on external datasets.}
\label{tab:TaskA-En-Biased-Results}
\end{table}

We used 5 epochs and 4e-6 learning rate because they gave the best results as shown in appendix \ref{Learning_Rates_Epochs}.

The official scores and leader-board ranks of the teams for subtask A (English) are shown in Table \ref{tab:leader-board-subtask-A-English}.
%%%%%%%%%%%%%%%%%% subtask A English Leader-Board Results %%%%%%%%%%
%%%%%%%%%%%%%%%%%%%%%%%%%%%%%%%%%%%%%%%%%%%%%%%%%%%%%%%%%%%%%%%%%
\begin{table}[h]
\centering
\resizebox{\columnwidth}{!}{
\begin{tabular}{|c|l|c|}
\hline
\rowcolor[HTML]{739CB9} 
{\color[HTML]{FFFFFF} \textbf{Rank}} & {\color[HTML]{FFFFFF} \textbf{User}}                                  & \multicolumn{1}{l|}{\cellcolor[HTML]{739CB9}{\color[HTML]{FFFFFF} \textbf{F-1 sarcastic}}} \\ \hline
\rowcolor[HTML]{FFFFFF} 
1                                    & stce                                                                  & 0.6052                                                                                     \\ \hline
\rowcolor[HTML]{FFFFFF} 
2                                    & emma                                                                  & 0.5691                                                                                     \\ \hline
\rowcolor[HTML]{FFFFFF} 
3                                    & saroyehun                                                             & 0.5295                                                                                     \\ \hline
\rowcolor[HTML]{FFFFFF} 
\textbf{4}                           & \textbf{\begin{tabular}[c]{@{}l@{}}ShubhamKumarNigam\end{tabular}} & \textbf{0.4769}                                                                            \\ \hline
\end{tabular}
}
\caption{Scores and leader-board ranks for subtask A (English)}
\label{tab:leader-board-subtask-A-English}
\end{table}

\textbf{ELECTRA:}
We used the ELECTRA model on the external datasets with Type-II preprocessing, 6e-6 learning rate, and 5 epochs as they were performing the best as shown in Table \ref{tab:TaskA-En-ELECTRA}.

\begin{table}[h]
\centering
\begin{tabular}{|c|
>{\columncolor[HTML]{FFFFFF}}c |
>{\columncolor[HTML]{FFFFFF}}c |}
\hline
\cellcolor[HTML]{739CB9}{\color[HTML]{FFFFFF} \textbf{Biased}} &
  \cellcolor[HTML]{739CB9}{\color[HTML]{FFFFFF} \textbf{Val}} &
  \cellcolor[HTML]{739CB9}{\color[HTML]{FFFFFF} \textbf{Test}} \\ \hline
B0                                                           & 0.5525                         & 0.4684                         \\ \hline
B1                                                           & 0.4002                         & 0.25                           \\ \hline
B2                                                           & 0.5738                         & 0.4762                         \\ \hline
B3                                                           & 0.4002                         & 0.25                           \\ \hline
B4                                                           & \cellcolor[HTML]{D5AB6D}0.5756 & \cellcolor[HTML]{6D9EEB}0.4879 \\ \hline
B5                                                           & 0.4002                         & 0.25                           \\ \hline
B6                                                           & 0.5468                         & 0.4642                         \\ \hline
B7                                                           & 0.5702                         & 0.4789                         \\ \hline
B8                                                           & 0.479                          & 0.5073                         \\ \hline
B9                                                           & 0.4002                         & 0.25                           \\ \hline
\end{tabular}
\caption{Results of the ELECTRA model for subtask A (English) on external datasets.}
\label{tab:TaskA-En-ELECTRA}
\end{table}

\textbf{BERT+BiLSTM with and without attention:}
The results we got using this architecture was not much stable (i.e. they may change when re-running the experiment) and thus we did not use this model for the official submission. More details about the results of the model can be found in the appendix \ref{BERT-BiLSTM-appendix}. 

%%%%%%%%%%%%%%%%%%%%%%%%%%%%%%%%%%%%%%%%%%%%%%
%%%%%%%%%%%%%%%%%% subtask A Arabic %%%%%%%%%%%%%
%%%%%%%%%%%%%%%%%%%%%%%%%%%%%%%%%%%%%%%%%%%%%%

\subsection{Subtask A (Arabic)}

\subsubsection{Datasets}
\textbf{Original:}
The training split of the original dataset for Arabic in Table \ref{tab:TaskA-Datasets} is splitted into training and validation datasets as shown in Table \ref{tab:TaskA-Ar-Original-Datasets}.

%%%%%%% TaskA-Ar-Original-Datasets %%%%%%%%%%%%%%%

\begin{table}[h]
\centering
\resizebox{0.8\columnwidth}{!}{
\begin{tabular}{|c|
>{\columncolor[HTML]{FFFFFF}}c |
>{\columncolor[HTML]{FFFFFF}}c |
>{\columncolor[HTML]{FFFFFF}}c |}
\hline
\cellcolor[HTML]{739CB9}{\color[HTML]{FFFFFF} \textbf{Dataset}} &
  \cellcolor[HTML]{739CB9}{\color[HTML]{FFFFFF} \textbf{Total}} &
  \cellcolor[HTML]{739CB9}{\color[HTML]{FFFFFF} \textbf{S\%}} &
  \cellcolor[HTML]{739CB9}{\color[HTML]{FFFFFF} \textbf{NS\%}} \\ \hline
Original Train &
  1861 &
  24 &
  76 \\ \hline
Original Val &
  1241 &
  24 &
  76 \\ \hline
\end{tabular}
}
\caption{Original datasets for subtask A (Arabic).}
\label{tab:TaskA-Ar-Original-Datasets}
\end{table}

%%%%%%%%%%%%%%%%%%%%%%%%%%%%%%%%%%%%%%%%%%

\textbf{External:}
Same as in subtask A (English), we created datasets which are biased towards the sarcastic class as shown in Table
\ref{tab:TaskA-Ar-External-Biased-Datasets}.

%%%%%%%%%%%% Biased external datasets for subtask A (Arabic) %%%%%%%%%%%%
%%%%%%%%%%%%%%%%%%%%%%%%%%%%%%%%%%%%%%%%%%%%%%%%%%%%%%%%%%%%%%%%%%%%%%%

\begin{table}[h]
\resizebox{\columnwidth}{!}{
\begin{tabular}{|
>{\columncolor[HTML]{FFFFFF}}c |c|
>{\columncolor[HTML]{FFFFFF}}c |
>{\columncolor[HTML]{FFFFFF}}c |
>{\columncolor[HTML]{FFFFFF}}c |
>{\columncolor[HTML]{FFFFFF}}c |}
\hline
\cellcolor[HTML]{739CB9}{\color[HTML]{FFFFFF} \textbf{Dataset}} &
  \cellcolor[HTML]{739CB9}{\color[HTML]{FFFFFF} \textbf{\begin{tabular}[c]{@{}c@{}}Contributing\\ Datasets\end{tabular}}} &
  \cellcolor[HTML]{739CB9}{\color[HTML]{FFFFFF} \textbf{\begin{tabular}[c]{@{}c@{}}Additional Non\\ Sarcastic Tweets\\ from SemEval\\ 2018-Training\end{tabular}}} &
  \cellcolor[HTML]{739CB9}{\color[HTML]{FFFFFF} \textbf{Total}} &
  \cellcolor[HTML]{739CB9}{\color[HTML]{FFFFFF} \textbf{S\%}} &
  \cellcolor[HTML]{739CB9}{\color[HTML]{FFFFFF} \textbf{NS\%}} \\ \hline
B0 &  & 0    & 4850 & 71 & 29 \\ \cline{1-1} \cline{3-6} 
B1 &  & 202  & 5052 & 68 & 32 \\ \cline{1-1} \cline{3-6} 
B2 &  & 404  & 5254 & 65 & 35 \\ \cline{1-1} \cline{3-6} 
B3 &  & 606  & 5456 & 63 & 37 \\ \cline{1-1} \cline{3-6} 
B4 &  & 808  & 5658 & 61 & 39 \\ \cline{1-1} \cline{3-6} 
B5 &  & 1010 & 5860 & 59 & 41 \\ \cline{1-1} \cline{3-6} 
B6 &  & 1212 & 6062 & 57 & 43 \\ \cline{1-1} \cline{3-6} 
B7 &  & 1414 & 6264 & 55 & 45 \\ \cline{1-1} \cline{3-6} 
B8 &  & 1616 & 6466 & 53 & 47 \\ \cline{1-1} \cline{3-6} 
B9 &
  \multirow{-10}{*}{\begin{tabular}[c]{@{}c@{}}Original Train+\\ ArSarcasm-v2 \\ Train\\ (only sarcastic)+\\ ArSarcasm-v2 \\ Test\\ (821 sarcastic)\end{tabular}} &
  1818 &
  6668 &
  52 &
  48 \\ \hline
\end{tabular}
}
\caption{External datasets for subtask A (Arabic)}
\label{tab:TaskA-Ar-External-Biased-Datasets}
\end{table}

%%%%%%%%%%%%%%%%%%%%%%%%%%%%%%%%%%%%%%%%%%%%%%%%%%

\subsubsection{Approaches}
The approached used here are similar to subtask A (English) except for the used transformers. Since the data is in the Arabic language, we tried some models from The \textbf{C}omputational \textbf{A}pproaches to \textbf{M}od\textbf{e}ling \textbf{L}anguage (CAMeL) research lab \footnote{\href{https://huggingface.co/CAMeL-Lab}{HuggingFace CAMeL-Lab}}. They majorly focused on Arabic and Arabic dialect processing, machine translation, text analysis, and dialogue systems. 

The models are available on the Hugging Face library. CAMeLBERT is a collection of BERT models pre-trained on Arabic texts with different sizes and variants \cite{inoue-etal-2021-interplay}. They released pre-trained language models for Modern Standard Arabic (MSA), dialectal Arabic (DA), and classical Arabic (CA). We tried CAMeLBERT-DA and CAMeLBERT-Mix for sarcasm detection. Likewise, we tried the AraBERT v2 which is a pre-trained BERT based on Google's BERT architecture for Arabic Language Understanding\footnote{\href{https://huggingface.co/aubmindlab/bert-base-arabertv02}{HuggingFace Bert-Base-Arabert-v02}} \cite{antoun2020arabert}.

\subsubsection{Results and Analysis}
Metric: The main metric is F1-score for the sarcastic class.

\textbf{BERT:}
We used CAMeLBERT-Mix as it performed the best among other BERT models. We applied it on the external datasets with non-weighted cross entropy loss function, 5 epochs, and 2e-5 learning rate because they gave the best results which are shown in Table \ref{tab:TaskA-Ar-Biased-Results}.

\begin{table}[h]
\centering
\begin{tabular}{|c|
>{\columncolor[HTML]{FFFFFF}}c |
>{\columncolor[HTML]{FFFFFF}}c |}
\hline
\cellcolor[HTML]{739CB9}{\color[HTML]{FFFFFF} \textbf{Biased}} &
  \cellcolor[HTML]{739CB9}{\color[HTML]{FFFFFF} \textbf{Val}} &
  \cellcolor[HTML]{739CB9}{\color[HTML]{FFFFFF} \textbf{Test}} \\ \hline
B0 & 0.7168 & 0.3438 \\ \hline
B1 & 0.7025 & 0.4163 \\ \hline
B2 &
  \cellcolor[HTML]{D5AB6D}0.7131 &
  \cellcolor[HTML]{6AA84F}0.4071 \\ \hline
B3 & 0.6804 & 0.4335 \\ \hline
B4 & 0.7015 & 0.4186 \\ \hline
B5 & 0.6927 & 0.4332 \\ \hline
B6 & 0.7012 & 0.4048 \\ \hline
B7 & 0.7124 & 0.4365 \\ \hline
B8 & 0.6731 & 0.4589 \\ \hline
B9 & 0.7094 & 0.4589 \\ \hline
\end{tabular}
\caption{Results of the BERT model for subtask A
(Arabic) on external datasets.}
\label{tab:TaskA-Ar-Biased-Results}
\end{table}

\textbf{BERT+BiLSTM+Attention:}
Same in subtask A (English), We used attention with BiLSTM on top of BERT model. The results also were not much stable. However, the best results for this architecture occurred when using 5 epochs and 9e-6 learning rate on B3 and B9 datasets as shown in Table \ref{tab:TaskA-Ar-BERT-RNN-ATT}. 

\begin{table}[h]
\centering
\begin{tabular}{|c|c|
>{\columncolor[HTML]{FFFFFF}}c |
>{\columncolor[HTML]{6D9EEB}}c |}
\hline
\cellcolor[HTML]{739CB9}{\color[HTML]{FFFFFF} \textbf{Biased}} &
  \cellcolor[HTML]{739CB9}{\color[HTML]{FFFFFF} \textbf{\begin{tabular}[c]{@{}c@{}}Hidden\\ State Size\end{tabular}}} &
  \cellcolor[HTML]{739CB9}{\color[HTML]{FFFFFF} \textbf{Val}} &
  \cellcolor[HTML]{739CB9}{\color[HTML]{FFFFFF} \textbf{Test}} \\ \hline
B3 &
  50 &
  0.6849 &
  0.4234 \\ \hline
B9 &
  1000 &
  0.7123 &
  0.4693 \\ \hline
\end{tabular}
\caption{Results of the BERT+BiLSTM+Attention model for subtask A (Arabic)}
\label{tab:TaskA-Ar-BERT-RNN-ATT}
\end{table}

%%%%%%%%%%%%%%%%%% subtask A Arabic Leader-Board Results %%%%%%%%%%
%%%%%%%%%%%%%%%%%%%%%%%%%%%%%%%%%%%%%%%%%%%%%%%%%%%%%%%%%%%%%%%%%

The official scores and leader-board ranks of the teams for subtask A (Arabic) are shown in Table \ref{tab:leader-board-subtask-A-Arabic}.

\begin{table}[h]
\centering
\resizebox{\columnwidth}{!}{
\begin{tabular}{|c|l|c|}
\hline
\rowcolor[HTML]{648CA8} 
{\color[HTML]{FFFFFF} \textbf{Rank}} & {\color[HTML]{FFFFFF} \textbf{User}}                                  & \multicolumn{1}{l|}{\cellcolor[HTML]{648CA8}{\color[HTML]{FFFFFF} \textbf{F-1 sarcastic}}} \\ \hline
\rowcolor[HTML]{FFFFFF} 
1                                    & Abdelkader                                                            & 0.5632                                                                                     \\ \hline
\rowcolor[HTML]{FFFFFF} 
2                                    & Aya                                                                   & 0.5076                                                                                     \\ \hline
\rowcolor[HTML]{FFFFFF} 
3                                    & rematchka                                                             & 0.4767                                                                                     \\ \hline
\rowcolor[HTML]{FFFFFF} 
\textbf{10}                          & \textbf{\begin{tabular}[c]{@{}l@{}}ShubhamKumarNigam\end{tabular}} & \textbf{0.4072}                                                                            \\ \hline
\end{tabular}
}
\caption{Scores and leader-board ranks for subtask A
(Arabic)}
\label{tab:leader-board-subtask-A-Arabic}
\end{table}

%%%%%%%%%%%%%%%%%%%%%%%%%%%%%%%%%%%%%%%%%%%%%%
%%%%%%%%%%%%%%%%%%% subtask B %%%%%%%%%%%%%%%%%%%
%%%%%%%%%%%%%%%%%%%%%%%%%%%%%%%%%%%%%%%%%%%%%%

\subsection{Subtask B}

\subsubsection{Datasets}
\textbf{Original:}
The training split of the original dataset for English in Table \ref{tab:TaskB-Datasets} is splitted into training and validation datasets as shown in Table \ref{tab:TaskB-En-Original-Datasets}.

%%%%%%%%%%%%%%%%%%%%%%%%%%%%%%%%%%%%%%

\begin{table}[h]
\resizebox{\columnwidth}{!}{
\begin{tabular}{|c|
>{\columncolor[HTML]{FFFFFF}}c |
>{\columncolor[HTML]{FFFFFF}}c |
>{\columncolor[HTML]{FFFFFF}}c |
>{\columncolor[HTML]{FFFFFF}}c |}
\hline
\cellcolor[HTML]{739CB9}{\color[HTML]{FFFFFF} } &
  \cellcolor[HTML]{739CB9}{\color[HTML]{FFFFFF} } &
  \cellcolor[HTML]{739CB9}{\color[HTML]{FFFFFF} \textbf{Sarcasm}} &
  \cellcolor[HTML]{739CB9}{\color[HTML]{FFFFFF} \textbf{Irony}} &
  \cellcolor[HTML]{739CB9}{\color[HTML]{FFFFFF} \textbf{Satire}} \\ \cline{3-5} 
\multirow{-2}{*}{\cellcolor[HTML]{739CB9}{\color[HTML]{FFFFFF} \textbf{Dataset}}} &
  \multirow{-2}{*}{\cellcolor[HTML]{739CB9}{\color[HTML]{FFFFFF} \textbf{Total}}} &
  \cellcolor[HTML]{739CB9}{\color[HTML]{FFFFFF} \textbf{\begin{tabular}[c]{@{}c@{}}Under-\\ statement\end{tabular}}} &
  \cellcolor[HTML]{739CB9}{\color[HTML]{FFFFFF} \textbf{\begin{tabular}[c]{@{}c@{}}Over-\\ statement\end{tabular}}} &
  \cellcolor[HTML]{739CB9}{\color[HTML]{FFFFFF} \textbf{\begin{tabular}[c]{@{}c@{}}Rhetorical\\ question\end{tabular}}} \\ \hline
                                                                           & \cellcolor[HTML]{FFFFFF}                      & 67.60\% & 15.10\% & 2.60\%  \\ \cline{3-5} 
\multirow{-2}{*}{\begin{tabular}[c]{@{}c@{}}Original\\ Train\end{tabular}} & \multirow{-2}{*}{\cellcolor[HTML]{FFFFFF}606} & 1\%     & 3.50\%  & 10.20\% \\ \hline
                                                                           & \cellcolor[HTML]{FFFFFF}                      & 70\%    & 14.20\% & 1.90\%  \\ \cline{3-5} 
\multirow{-2}{*}{\begin{tabular}[c]{@{}c@{}}Original\\ Val\end{tabular}}   & \multirow{-2}{*}{\cellcolor[HTML]{FFFFFF}261} & 1\%     & 4.50\%  & 8.40\%  \\ \hline
\end{tabular}
}
\caption{Original datasets for subtask B (English)}
\label{tab:TaskB-En-Original-Datasets}
\end{table}

%%%%%%%%%%%%%%%%%%%%%%%%%%%%%%%%%%%%%%%%%

\textbf{External:}
The original and external datasets presented in Table \ref{tab:TaskB-Datasets} (without the validation dataset) were added together to form a new dataset (Ext-NB). Then the resulting dataset was balanced either by using word embeddings (Ext-UW) or by repeating instances (Ext-UR). We created a dataset (Ext-EB) to give more importance to the labels of low number of instances by repeating the instances of these labels up to the limits specified by these heuristic formulas:\\
{\small \#irony=\#sarcasm*(1+1/sqrt(\#irony)) \hfill (1)}
{\small \#satire=\#sarcasm*(1+2/sqrt(\#satire))\hfill (2)}\\
{\small \#understatement=\#sarcasm*(1+3/sqrt(understatement)) \hfill (3)}\\
{\small \#overstatement=\#sarcasm*(1+1.5/sqrt(overstatement)) \hfill (4)}\\
{\small \#rhetorical=\#sarcasm*(1+1.2/sqrt(\#rhetorical)) \hfill (5)}\\
The datasets are shown in Table \ref{tab:TaskB-En-External-Datasets}.

%%%%%%%%%%%%%% External datasets for subtask A (English) %%%%%%%%%%%%
%%%%%%%%%%%%%%%%%%%%%%%%%%%%%%%%%%%%%%%%%%%%%%%%%%%%%%%%%%%%%%%%%%

\begin{table}[h]
\resizebox{\columnwidth}{!}{
\begin{tabular}{|c|c|c|
>{\columncolor[HTML]{FFFFFF}}c |c|
>{\columncolor[HTML]{FFFFFF}}c |}
\hline
\cellcolor[HTML]{739CB9}{\color[HTML]{FFFFFF} } &
  \cellcolor[HTML]{739CB9}{\color[HTML]{FFFFFF} } &
  \cellcolor[HTML]{739CB9}{\color[HTML]{FFFFFF} } &
  \cellcolor[HTML]{739CB9}{\color[HTML]{FFFFFF} } &
  \cellcolor[HTML]{739CB9}{\color[HTML]{FFFFFF} \textbf{Sarcasm}} &
  \cellcolor[HTML]{739CB9}{\color[HTML]{FFFFFF} \textbf{\begin{tabular}[c]{@{}c@{}}Under-\\ statement\end{tabular}}} \\ \cline{5-6} 
\cellcolor[HTML]{739CB9}{\color[HTML]{FFFFFF} } &
  \cellcolor[HTML]{739CB9}{\color[HTML]{FFFFFF} } &
  \cellcolor[HTML]{739CB9}{\color[HTML]{FFFFFF} } &
  \cellcolor[HTML]{739CB9}{\color[HTML]{FFFFFF} } &
  \cellcolor[HTML]{739CB9}{\color[HTML]{FFFFFF} \textbf{Irony}} &
  \cellcolor[HTML]{739CB9}{\color[HTML]{FFFFFF} \textbf{\begin{tabular}[c]{@{}c@{}}Over-\\ statement\end{tabular}}} \\ \cline{5-6} 
\multirow{-5}{*}{\cellcolor[HTML]{739CB9}{\color[HTML]{FFFFFF} \textbf{Dataset}}} &
  \multirow{-5}{*}{\cellcolor[HTML]{739CB9}{\color[HTML]{FFFFFF} \textbf{Balanced}}} &
  \multirow{-5}{*}{\cellcolor[HTML]{739CB9}{\color[HTML]{FFFFFF} \textbf{\begin{tabular}[c]{@{}c@{}}Contributing\\ Datasets\end{tabular}}}} &
  \multirow{-5}{*}{\cellcolor[HTML]{739CB9}{\color[HTML]{FFFFFF} \textbf{Total}}} &
  \cellcolor[HTML]{739CB9}{\color[HTML]{FFFFFF} \textbf{Satire}} &
  \cellcolor[HTML]{739CB9}{\color[HTML]{FFFFFF} \textbf{\begin{tabular}[c]{@{}c@{}}Rhetorical\\ question\end{tabular}}} \\ \hline
\cellcolor[HTML]{FFFFFF} &
  \cellcolor[HTML]{FFFFFF} &
   &
  \cellcolor[HTML]{FFFFFF} &
  \cellcolor[HTML]{FFFFFF}55.90\% &
  1.20\% \\ \cline{5-6} 
\cellcolor[HTML]{FFFFFF} &
  \cellcolor[HTML]{FFFFFF} &
   &
  \cellcolor[HTML]{FFFFFF} &
  \cellcolor[HTML]{FFFFFF}22.30\% &
  5.50\% \\ \cline{5-6} 
\multirow{-3}{*}{\cellcolor[HTML]{FFFFFF}Ext-NB} &
  \multirow{-3}{*}{\cellcolor[HTML]{FFFFFF}Not Balanced} &
   &
  \multirow{-3}{*}{\cellcolor[HTML]{FFFFFF}1203} &
  \cellcolor[HTML]{FFFFFF}6.40\% &
  8.70\% \\ \cline{1-2} \cline{4-6} 
 &
   &
   &
  \cellcolor[HTML]{FFFFFF} &
  16.50\% &
  16.60\% \\ \cline{5-6} 
 &
   &
   &
  \cellcolor[HTML]{FFFFFF} &
  \cellcolor[HTML]{FFFFFF}16.50\% &
  16.80\% \\ \cline{5-6} 
\multirow{-3}{*}{Ext-UW} &
  \multirow{-3}{*}{\begin{tabular}[c]{@{}c@{}}Using Word\\ Embedding\end{tabular}} &
   &
  \multirow{-3}{*}{\cellcolor[HTML]{FFFFFF}4336} &
  \cellcolor[HTML]{FFFFFF}16.60\% &
  17\% \\ \cline{1-2} \cline{4-6} 
 &
   &
   &
  \cellcolor[HTML]{FFFFFF} &
  16.50\% &
  16.60\% \\ \cline{5-6} 
 &
   &
   &
  \cellcolor[HTML]{FFFFFF} &
  \cellcolor[HTML]{FFFFFF}16.50\% &
  16.80\% \\ \cline{5-6} 
\multirow{-3}{*}{Ext-UR} &
  \multirow{-3}{*}{\begin{tabular}[c]{@{}c@{}}Using\\ Repetition\end{tabular}} &
   &
  \multirow{-3}{*}{\cellcolor[HTML]{FFFFFF}4336} &
  \cellcolor[HTML]{FFFFFF}16.60\% &
  16.90\% \\ \cline{1-2} \cline{4-6} 
 &
  \cellcolor[HTML]{FFFFFF} &
   &
  \cellcolor[HTML]{FFFFFF} &
  15\% &
  18.80\% \\ \cline{5-6} 
 &
  \cellcolor[HTML]{FFFFFF} &
   &
  \cellcolor[HTML]{FFFFFF} &
  \cellcolor[HTML]{FFFFFF}15.80\% &
  17\% \\ \cline{5-6} 
\multirow{-3}{*}{Ext-EB} &
  \multirow{-3}{*}{\cellcolor[HTML]{FFFFFF}Not Balanced} &
  \multirow{-12}{*}{\begin{tabular}[c]{@{}c@{}}Original\\ Train +\\ Twitter API\\ Train +\\ Twitter API\\ Test\end{tabular}} &
  \multirow{-3}{*}{\cellcolor[HTML]{FFFFFF}5314} &
  \cellcolor[HTML]{FFFFFF}16.70\% &
  16.80\% \\ \hline
\end{tabular}
}
\caption{External datasets for subtask B (English).}
\label{tab:TaskB-En-External-Datasets}
\end{table}

%%%%%%%%%%%%%%%%%%%%%%%%%%%%%%%%%%%%%%%%%%%%%%%%%%%%%%%%%%%%%%%%%%

\subsubsection{Approaches}
This subtask primarily focused on BERTweet-large. As it is a multi labeling subtask, we used sigmoid as the activation function in the last layer and binary cross entropy as the loss function.

\subsubsection{Results and Analysis}
Metric: The main metric is Macro-F1 score.

\textbf{BERT:}
We used BERTweet-large on the external datasets using 5 epochs, 6e-6 learning rate, and Type-II preprocessing. The results are shown in Table \ref{tab:TaskB-En-External-Results}.

\begin{table}[h]
\centering
\begin{tabular}{|c|
>{\columncolor[HTML]{FFFFFF}}c |
>{\columncolor[HTML]{FFFFFF}}c |}
\hline
\cellcolor[HTML]{739CB9}{\color[HTML]{FFFFFF} \textbf{Dataset}} &
  \cellcolor[HTML]{739CB9}{\color[HTML]{FFFFFF} \textbf{Val}} &
  \cellcolor[HTML]{739CB9}{\color[HTML]{FFFFFF} \textbf{Test}} \\ \hline
\cellcolor[HTML]{FFFFFF}Ext-NB & 0.1513                         & 0.038                          \\ \hline
Ext-UW                         & 0.318                          & 0.0716                         \\ \hline
Ext-UR                         & 0.3412                         & 0.076                          \\ \hline
Ext-EB                         & \cellcolor[HTML]{D5AB6D}0.4152 & \cellcolor[HTML]{6AA84F}0.0778 \\ \hline
\end{tabular}
\caption{Results of the BERT model for subtask B
(English) on external datasets.}
\label{tab:TaskB-En-External-Results}
\end{table}

%%%%%%%%%%%%%%%%%% subtask B Leader-Board Results %%%%%%%%%%%%%%%%%%
%%%%%%%%%%%%%%%%%%%%%%%%%%%%%%%%%%%%%%%%%%%%%%%%%%%%%%%%%%%%%%%%%
The official scores and leader-board ranks of the teams for subtask B (English) are shown in Table \ref{tab:leader-board-subtask-B-English}.

\begin{table}[h]
\centering
\resizebox{\columnwidth}{!}{
\begin{tabular}{|c|l|r|}
\hline
\rowcolor[HTML]{648CA8} 
{\color[HTML]{FFFFFF} \textbf{Rank}} & {\color[HTML]{FFFFFF} \textbf{User}}                                  & \multicolumn{1}{l|}{\cellcolor[HTML]{648CA8}{\color[HTML]{FFFFFF} \textbf{macro F-score}}} \\ \hline
\rowcolor[HTML]{FFFFFF} 
1                                    & Duxy                                                                  & 0.163                                                                                      \\ \hline
\rowcolor[HTML]{FFFFFF} 
2                                    & Abdelkader                                                            & 0.0875                                                                                     \\ \hline
\rowcolor[HTML]{FFFFFF} 
3                                    & robvanderg                                                            & 0.0851                                                                                     \\ \hline
\rowcolor[HTML]{FFFFFF} 
\textbf{6}                           & \textbf{\begin{tabular}[c]{@{}l@{}}ShubhamKumarNigam\end{tabular}} & \textbf{0.0778}                                                                            \\ \hline
\end{tabular}
}
\caption{Scores and leader-board ranks for subtask B (English)}
\label{tab:leader-board-subtask-B-English}
\end{table}
%%%%%%%%%%%%%%%%%%%%%%%%%%%%%%%%%%%%%%%%%%%%%%
%%%%%%%%%%%%%%%% subtask C English %%%%%%%%%%%%%%
%%%%%%%%%%%%%%%%%%%%%%%%%%%%%%%%%%%%%%%%%%%%%%
\subsection{Subtask C (English)}

\subsubsection{Datasets}
\textbf{Original:}
The training split of the original dataset for English in Table \ref{tab:TaskA-Datasets} is splitted into training and validation datasets. As we do not have external datasets here, so we augmented the training split once with word embeddings (Original-Embedding) and once with repetition (Original-Repetition). We swapped between the tweet and its rephrase for half of the instances together with flipping the value of the sarcastic field, so that the model will be able to learn. Otherwise, it may always predict the first text as the tweet and the second one as the rephrase. The datasets are shown in Table \ref{tab:TaskC-En-Original-Datasets}.

\begin{table}[h]
\centering
\begin{tabular}{|c|
>{\columncolor[HTML]{FFFFFF}}c |}
\hline
\cellcolor[HTML]{739CB9}{\color[HTML]{FFFFFF} \textbf{Dataset}} & \cellcolor[HTML]{739CB9}{\color[HTML]{FFFFFF} \textbf{Total}} \\ \hline
Original-Train   & 606  \\ \hline
Original-Validation & 261  \\ \hline
Original-Embedding  & 1606 \\ \hline
Original-Repetition & 1606 \\ \hline
\end{tabular}
\caption{Original datasets for subtask C (English)}
\label{tab:TaskC-En-Original-Datasets}
\end{table}

\subsubsection{Approaches}
Organizers provide a sarcastic text, and its non-sarcastic rephrase, i.e., two texts convey the same meaning. Since the input format changed in this subtask, we input both texts together to the BERT model as one text separating them by the separating token. We focused on transformers which are trained on question-answering tasks. We tried BERT models trained on the \textbf{S}tanford \textbf{Qu}estion \textbf{A}nswering \textbf{D}ataset (SQuAD) dataset \cite{rajpurkar-etal-2018-know}. SQuAD is a reading comprehension dataset consisting of questions posed by crowdworkers on a set of Wikipedia articles.

We took models from the Hugging Face library; one is BERT large model (cased)\footnote{\href{https://huggingface.co/bert-large-cased-whole-word-masking-finetuned-squad}{HuggingFace Bert-Large-Cased-Whole-Word-Masking-Finetuned-Squad}}, trained on whole word masking, and fine-tuned on the SQuAD dataset. Another is BERT base model (uncased)\footnote{\href{https://huggingface.co/twmkn9/bert-base-uncased-squad2}{HuggingFace Bert-Base-Uncased-Squad2}}, trained on Masked language modeling (MLM), and fine-tuned on the SQuAD dataset. 

\subsubsection{Results and Analysis}
Metric: The main metric is the accuracy.

\textbf{BERT:}
We used BERT large model (cased) on the original datasets, because it gave better results among the other models, using 15 epochs, 8e-6 learning rate, and the cross entropy as the loss function. The results are shown in Table \ref{tab:TaskC-En-Original-Results}. We did our submission using Type-I preprocessing even though Type-II was performing better on the validation dataset.

\begin{table}[h]
\centering
\begin{tabular}{|c|c|
>{\columncolor[HTML]{FFFFFF}}c |
>{\columncolor[HTML]{FFFFFF}}c |}
\hline
\cellcolor[HTML]{739CB9}{\color[HTML]{FFFFFF} \textbf{Dataset}} &
  \cellcolor[HTML]{739CB9}{\color[HTML]{FFFFFF} \textbf{Type}} &
  \cellcolor[HTML]{739CB9}{\color[HTML]{FFFFFF} \textbf{Val}} &
  \cellcolor[HTML]{739CB9}{\color[HTML]{FFFFFF} \textbf{Test}} \\ \hline
 &
  \cellcolor[HTML]{FFFFFF}I &
  \cellcolor[HTML]{6D9EEB}0.951 &
  \cellcolor[HTML]{6AA84F}0.79 \\ \cline{2-4} 
 &
  \cellcolor[HTML]{FFFFFF}II &
  0.9395 &
  0.8 \\ \cline{2-4} 
 &
  \cellcolor[HTML]{FFFFFF}III &
  0.9193 &
  0.83 \\ \cline{2-4} 
\multirow{-4}{*}{\begin{tabular}[c]{@{}c@{}}Original-\\ Train\end{tabular}} &
  \cellcolor[HTML]{FFFFFF}IV &
  0.9135 &
  0.79 \\ \hline
 &
  I &
  0.9654 &
  0.83 \\ \cline{2-4} 
 &
  \cellcolor[HTML]{FFFFFF}II &
  0.9654 &
  0.8 \\ \cline{2-4} 
 &
  III &
  0.9366 &
  0.82 \\ \cline{2-4} 
\multirow{-4}{*}{\begin{tabular}[c]{@{}c@{}}Original-\\ Embedding\end{tabular}} &
  IV &
  0.8588 &
  0.72 \\ \hline
 &
  I &
  0.9395 &
  0.8 \\ \cline{2-4} 
 &
  \cellcolor[HTML]{FFFFFF}II &
  0.9222 &
  0.815 \\ \cline{2-4} 
 &
  III &
  0.9078 &
  0.8 \\ \cline{2-4} 
\multirow{-4}{*}{\begin{tabular}[c]{@{}c@{}}Original-\\ Repetition\end{tabular}} &
  IV &
  0.9078 &
  0.68 \\ \hline
\end{tabular}
\caption{Results of the BERT model for subtask C (English) on original datasets.}
\label{tab:TaskC-En-Original-Results}
\end{table}

%%%%%%%%%%%%%%%%%% subtask C English Leader-Board Results %%%%%%%%%%
%%%%%%%%%%%%%%%%%%%%%%%%%%%%%%%%%%%%%%%%%%%%%%%%%%%%%%%%%%%%%%%%%
The official scores and leader-board ranks of the teams for subtask C (English) are shown in Table \ref{tab:leader-board-subtask-C-English}.

\begin{table}[h]
\centering
\resizebox{\columnwidth}{!}{
\begin{tabular}{|c|l|r|}
\hline
\rowcolor[HTML]{648CA8} 
{\color[HTML]{FFFFFF} \textbf{Rank}} &
  {\color[HTML]{FFFFFF} \textbf{User}} &
  \multicolumn{1}{l|}{\cellcolor[HTML]{648CA8}{\color[HTML]{FFFFFF} \textbf{Accuracy}}} \\ \hline
\rowcolor[HTML]{FFFFFF} 
1 & emma      & 0.87  \\ \hline
\rowcolor[HTML]{FFFFFF} 
2 & lizefeng  & 0.855 \\ \hline
\rowcolor[HTML]{FFFFFF} 
3 & leon14138 & 0.805 \\ \hline
\rowcolor[HTML]{FFFFFF} 
\textbf{4} &
  \textbf{\begin{tabular}[c]{@{}l@{}}ShubhamKumarNigam\end{tabular}} &
  \textbf{0.79} \\ \hline
\end{tabular}
}
\caption{Scores and leader-board ranks for subtask C
(English)}
\label{tab:leader-board-subtask-C-English}
\end{table}

%%%%%%%%%%%%%%%%%%%%%%%%%%%%%%%%%%%%%%%%%%%%%%
%%%%%%%%%%%%%%%% subtask C Arabic %%%%%%%%%%%%%%
%%%%%%%%%%%%%%%%%%%%%%%%%%%%%%%%%%%%%%%%%%%%%%

\subsection{Subtask C (Arabic)}

\subsubsection{Datasets}
\textbf{Original:}
The datasets were generated in the same way as in subtask C (Engligh) and are shown in Table \ref{tab:TaskC-Ar-Original-Datasets}

\begin{table}[h]
\centering
\begin{tabular}{|c|
>{\columncolor[HTML]{FFFFFF}}c |}
\hline
\cellcolor[HTML]{739CB9}{\color[HTML]{FFFFFF} \textbf{Dataset}} & \cellcolor[HTML]{739CB9}{\color[HTML]{FFFFFF} \textbf{Total}} \\ \hline
Original-Train   & 521  \\ \hline
Original-Validation & 224  \\ \hline
Original-Embedding  & 1521 \\ \hline
Original-Repetition & 1521 \\ \hline
\end{tabular}
\caption{Original datasets for subtask C (Arabic)}
\label{tab:TaskC-Ar-Original-Datasets}
\end{table}

\subsubsection{Approaches}
The approached used here are similar to subtask C (English) except for the used transformers. Since the data is in the Arabic language, we took a couple of models from the Hugging Face library like the multilingual model mBERT base (cased), trained on the QA dataset in seven languages and fine-tuned on the combination of XQuAD \cite{artetxe-etal-2020-cross} and MLQA \cite{lewis-etal-2020-mlqa} datasets. We compared their performance to the CAMeLBERT-Mix model.

\subsubsection{Results and Analysis}
Metric: The main metric is the accuracy.

\textbf{BERT:}
We used CAMeLBERT-Mix model on the original datasets using 5 epochs, 3e-5 learning rate, and the cross entropy as the loss function. The results are shown in Table \ref{tab:TaskC-Ar-Original-Results}. We did our submission using Type-II preprocessing as it is the best performing.

\begin{table}[h]
\centering
\begin{tabular}{|c|c|
>{\columncolor[HTML]{FFFFFF}}c |
>{\columncolor[HTML]{FFFFFF}}c |}
\hline
\cellcolor[HTML]{739CB9}{\color[HTML]{FFFFFF} \textbf{Dataset}} &
  \cellcolor[HTML]{739CB9}{\color[HTML]{FFFFFF} \textbf{Type}} &
  \cellcolor[HTML]{739CB9}{\color[HTML]{FFFFFF} \textbf{Val}} &
  \cellcolor[HTML]{739CB9}{\color[HTML]{FFFFFF} \textbf{Test}} \\ \hline
 &
  I &
  0.6242 &
  0.5 \\ \cline{2-4} 
 &
  \cellcolor[HTML]{FFFFFF}II &
  0.6242 &
  0.5 \\ \cline{2-4} 
 &
  III &
  0.6711 &
  0.72 \\ \cline{2-4} 
\multirow{-4}{*}{\begin{tabular}[c]{@{}c@{}}Original-\\ Train\end{tabular}} &
  IV &
  0.3758 &
  0.5 \\ \hline
 &
  I &
  0.8792 &
  0.825 \\ \cline{2-4} 
 &
  \cellcolor[HTML]{FFFFFF}II &
  0.8792 &
  0.845 \\ \cline{2-4} 
 &
  III &
  0.8691 &
  0.845 \\ \cline{2-4} 
\multirow{-4}{*}{\begin{tabular}[c]{@{}c@{}}Original-\\ Embedding\end{tabular}} &
  IV &
  0.7517 &
  0.71 \\ \hline
 &
  I &
  0.8993 &
  0.855 \\ \cline{2-4} 
 &
  \cellcolor[HTML]{FFFFFF}II &
  \cellcolor[HTML]{6D9EEB}0.9262 &
  \cellcolor[HTML]{6AA84F}0.87 \\ \cline{2-4} 
 &
  III &
  0.8658 &
  0.86 \\ \cline{2-4} 
\multirow{-4}{*}{\begin{tabular}[c]{@{}c@{}}Original-\\ Repetition\end{tabular}} &
  IV &
  0.6409 &
  0.705 \\ \hline
\end{tabular}
\caption{Results of BERT model for subtask C (Arabic) on original datasets}
\label{tab:TaskC-Ar-Original-Results}
\end{table}

%%%%%%%%%%%%%%%%%% subtask C Arabic Leader-Board Results %%%%%%%%%%
%%%%%%%%%%%%%%%%%%%%%%%%%%%%%%%%%%%%%%%%%%%%%%%%%%%%%%%%%%%%%%%%%
The official scores and leader-board ranks of the teams for subtask C (Arabic) is shown in Table \ref{tab:leader-board-subtask-C-Arabic}.

\begin{table}[h]
\centering
\resizebox{\columnwidth}{!}{
\begin{tabular}{|c|l|r|}
\hline
\rowcolor[HTML]{648CA8} 
{\color[HTML]{FFFFFF} \textbf{Rank}} &
  {\color[HTML]{FFFFFF} \textbf{User}} &
  \multicolumn{1}{l|}{\cellcolor[HTML]{648CA8}{\color[HTML]{FFFFFF} \textbf{Accuracy}}} \\ \hline
\rowcolor[HTML]{FFFFFF} 
1 & lizefeng      & 0.93  \\ \hline
\rowcolor[HTML]{FFFFFF} 
2 & AlamiHamza    & 0.885 \\ \hline
\rowcolor[HTML]{FFFFFF} 
3 & maryam.najafi & 0.875 \\ \hline
\rowcolor[HTML]{FFFFFF} 
\textbf{4} &
  \textbf{\begin{tabular}[c]{@{}l@{}}ShubhamKumarNigam\end{tabular}} &
  \textbf{0.87} \\ \hline
\end{tabular}
}
\caption{Scores and leader-board ranks for subtask C
(Arabic)}
\label{tab:leader-board-subtask-C-Arabic}
\end{table}

%Conclusion
\section{Conclusion and Future Direction}

We built our systems using stand-alone transformer based models and hierarchical ones by adding a BiLSTM layer with or without attention mechanism. In addition to this, we created new datasets depending on the subtask using a combination of original and external datasets besides augmenting using word embedding or repetition. Our results shows that the augmented datasets enhanced the results for all subtasks. Moreover, we found that the fine-tuned stand-alone transformer based models in each subtask gave the best results especially with Type-II preprocessing. We also showed the enhancement when using a weighted loss function and the effect of using different learning-rates, epochs, and preprocessing types. Moreover, the rank of our team was consistent across most of the subtasks (the fourth rank) which shows the robustness of the techniques we used.

\bibliography{anthology,custom}

\begin{thebibliography}{17}
\expandafter\ifx\csname natexlab\endcsname\relax\def\natexlab#1{#1}\fi

\bibitem[{Abu~Farha et~al.(2022)Abu~Farha, Oprea, Wilson, and
  Magdy}]{abufarha-etal-2022-semeval}
Ibrahim Abu~Farha, Silviu Oprea, Steven Wilson, and Walid Magdy. 2022.
\newblock {SemEval-2022 Task 6}: {iSarcasmEval}, {Intended Sarcasm Detection in
  English and Arabic}.
\newblock In \emph{Proceedings of the 16th International Workshop on Semantic
  Evaluation (SemEval-2022)}. Association for Computational Linguistics.

\bibitem[{Abu~Farha et~al.(2021)Abu~Farha, Zaghouani, and
  Magdy}]{abu-farha-etal-2021-overview}
Ibrahim Abu~Farha, Wajdi Zaghouani, and Walid Magdy. 2021.
\newblock \href {https://aclanthology.org/2021.wanlp-1.36} {Overview of the
  {WANLP} 2021 shared task on sarcasm and sentiment detection in {A}rabic}.
\newblock In \emph{Proceedings of the Sixth Arabic Natural Language Processing
  Workshop}, pages 296--305, Kyiv, Ukraine (Virtual). Association for
  Computational Linguistics.

\bibitem[{Antoun et~al.()Antoun, Baly, and Hajj}]{antoun2020arabert}
Wissam Antoun, Fady Baly, and Hazem Hajj.
\newblock Arabert: Transformer-based model for {A}rabic language understanding.
\newblock In \emph{LREC 2020 Workshop Language Resources and Evaluation
  Conference 11--16 May 2020}, page~9.

\bibitem[{Artetxe et~al.(2020)Artetxe, Ruder, and
  Yogatama}]{artetxe-etal-2020-cross}
Mikel Artetxe, Sebastian Ruder, and Dani Yogatama. 2020.
\newblock \href {https://doi.org/10.18653/v1/2020.acl-main.421} {On the
  cross-lingual transferability of monolingual representations}.
\newblock In \emph{Proceedings of the 58th Annual Meeting of the Association
  for Computational Linguistics}, pages 4623--4637, Online. Association for
  Computational Linguistics.

\bibitem[{Bing(2012)}]{bing2012sentiment}
L~Bing. 2012.
\newblock Sentiment analysis and opinion mining (synthesis lectures on human
  language technologies).
\newblock \emph{University of Illinois: Chicago, IL, USA}.

\bibitem[{Clark et~al.(2020)Clark, Luong, Le, and Manning}]{clark2020electra}
Kevin Clark, Minh-Thang Luong, Quoc~V. Le, and Christopher~D. Manning. 2020.
\newblock \href {https://openreview.net/pdf?id=r1xMH1BtvB} {{ELECTRA}:
  Pre-training text encoders as discriminators rather than generators}.
\newblock In \emph{ICLR}.

\bibitem[{Ghanem et~al.(2019)Ghanem, Karoui, Benamara, Moriceau, and
  Rosso}]{ghanem2019idat}
Bilal Ghanem, Jihen Karoui, Farah Benamara, V{\'e}ronique Moriceau, and Paolo
  Rosso. 2019.
\newblock Idat at fire2019: Overview of the track on irony detection in
  {A}rabic tweets.
\newblock In \emph{Proceedings of the 11th Forum for Information Retrieval
  Evaluation}, pages 10--13.

\bibitem[{Ghosh et~al.(2020)Ghosh, Vajpayee, and
  Muresan}]{ghosh-etal-2020-report}
Debanjan Ghosh, Avijit Vajpayee, and Smaranda Muresan. 2020.
\newblock \href {https://doi.org/10.18653/v1/2020.figlang-1.1} {A report on the
  2020 sarcasm detection shared task}.
\newblock In \emph{Proceedings of the Second Workshop on Figurative Language
  Processing}, pages 1--11, Online. Association for Computational Linguistics.

\bibitem[{Hee et~al.(2018)Hee, Lefever, and Hoste}]{VanHee2018}
Cynthia~Van Hee, Els Lefever, and Veronique Hoste. 2018.
\newblock \href {https://doi.org/10.18653/v1/s18-1005} {{SemEval}-2018 task 3:
  Irony detection in {English} tweets}.
\newblock In \emph{Proceedings of The 12th International Workshop on Semantic
  Evaluation}. Association for Computational Linguistics.

\bibitem[{Inoue et~al.(2021)Inoue, Alhafni, Baimukan, Bouamor, and
  Habash}]{inoue-etal-2021-interplay}
Go~Inoue, Bashar Alhafni, Nurpeiis Baimukan, Houda Bouamor, and Nizar Habash.
  2021.
\newblock The interplay of variant, size, and task type in {A}rabic pre-trained
  language models.
\newblock In \emph{Proceedings of the Sixth Arabic Natural Language Processing
  Workshop}, Kyiv, Ukraine (Online). Association for Computational Linguistics.

\bibitem[{Lewis et~al.(2020)Lewis, Oguz, Rinott, Riedel, and
  Schwenk}]{lewis-etal-2020-mlqa}
Patrick Lewis, Barlas Oguz, Ruty Rinott, Sebastian Riedel, and Holger Schwenk.
  2020.
\newblock \href {https://doi.org/10.18653/v1/2020.acl-main.653} {{MLQA}:
  Evaluating cross-lingual extractive question answering}.
\newblock In \emph{Proceedings of the 58th Annual Meeting of the Association
  for Computational Linguistics}, pages 7315--7330, Online. Association for
  Computational Linguistics.

\bibitem[{Maynard and Greenwood(2014)}]{maynard-greenwood-2014-cares}
Diana Maynard and Mark Greenwood. 2014.
\newblock \href
  {http://www.lrec-conf.org/proceedings/lrec2014/pdf/67_Paper.pdf} {Who cares
  about sarcastic tweets? investigating the impact of sarcasm on sentiment
  analysis.}
\newblock In \emph{Proceedings of the Ninth International Conference on
  Language Resources and Evaluation ({LREC}'14)}, pages 4238--4243, Reykjavik,
  Iceland. European Language Resources Association (ELRA).

\bibitem[{Nguyen et~al.(2020)Nguyen, Vu, and Nguyen}]{bertweet}
Dat~Quoc Nguyen, Thanh Vu, and Anh~Tuan Nguyen. 2020.
\newblock {BERT}weet: A pre-trained language model for {E}nglish tweets.
\newblock In \emph{Proceedings of the 2020 Conference on Empirical Methods in
  Natural Language Processing: System Demonstrations}, pages 9--14.

\bibitem[{Oprea and Magdy(2020)}]{oprea-magdy-2020-isarcasm}
Silviu Oprea and Walid Magdy. 2020.
\newblock \href {https://doi.org/10.18653/v1/2020.acl-main.118} {i{S}arcasm: A
  dataset of intended sarcasm}.
\newblock In \emph{Proceedings of the 58th Annual Meeting of the Association
  for Computational Linguistics}, pages 1279--1289, Online. Association for
  Computational Linguistics.

\bibitem[{Rajpurkar et~al.(2018)Rajpurkar, Jia, and
  Liang}]{rajpurkar-etal-2018-know}
Pranav Rajpurkar, Robin Jia, and Percy Liang. 2018.
\newblock \href {https://doi.org/10.18653/v1/P18-2124} {Know what you don{'}t
  know: Unanswerable questions for {SQ}u{AD}}.
\newblock In \emph{Proceedings of the 56th Annual Meeting of the Association
  for Computational Linguistics (Volume 2: Short Papers)}, pages 784--789,
  Melbourne, Australia. Association for Computational Linguistics.

\bibitem[{Rosenthal et~al.(2014)Rosenthal, Ritter, Nakov, and
  Stoyanov}]{rosenthal-etal-2014-semeval}
Sara Rosenthal, Alan Ritter, Preslav Nakov, and Veselin Stoyanov. 2014.
\newblock \href {https://doi.org/10.3115/v1/S14-2009} {{S}em{E}val-2014 task 9:
  Sentiment analysis in {T}witter}.
\newblock In \emph{Proceedings of the 8th International Workshop on Semantic
  Evaluation ({S}em{E}val 2014)}, pages 73--80, Dublin, Ireland. Association
  for Computational Linguistics.

\bibitem[{Van~Hee et~al.(2018)Van~Hee, Lefever, and
  Hoste}]{van-hee-etal-2018-semeval}
Cynthia Van~Hee, Els Lefever, and V{\'e}ronique Hoste. 2018.
\newblock \href {https://doi.org/10.18653/v1/S18-1005} {{S}em{E}val-2018 task
  3: Irony detection in {E}nglish tweets}.
\newblock In \emph{Proceedings of The 12th International Workshop on Semantic
  Evaluation}, pages 39--50, New Orleans, Louisiana. Association for
  Computational Linguistics.

\end{thebibliography}
\clearpage

\appendix
\section{Appendix}
\label{sec:appendix}

\subsection{Hierarchical Architecture}
\label{Hier_Network}
\begin{figure}[h]
\centering
\includegraphics[width=\linewidth ]{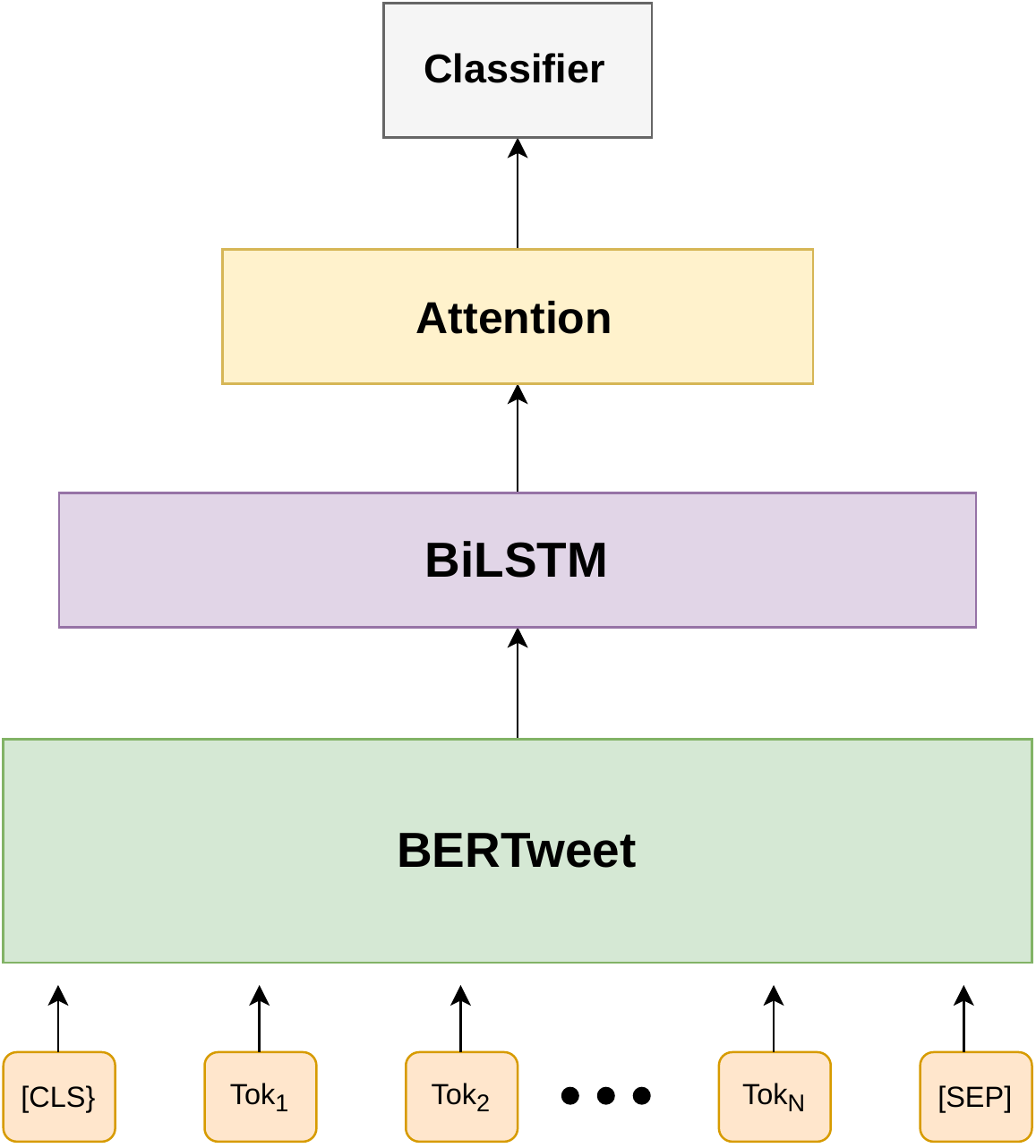}
\caption{Hierarchical architecture}
\label{fig:Hierarchical_Network}
\end{figure}

Figure \ref{fig:Hierarchical_Network} shows a hierarchical network based on a transformer. The input tokens are passed to the transformer (BERTweet), then the output token embeddings are passed to a Bi-LSTM layer which can be with or without attention mechanism.
%%%%%%% Hierarchical Network %%%%%

%%%%%%%%%%%%%%%%%%%%%%%%%%%%%%%%%%%%%%%%%%%%%%%%%%%

\subsection{Sarcasm Types Description}
\label{ssec:Sarcasm-types-description}
    \begin{enumerate}
        \item \textbf{Sarcasm:} tweets that contradict the state of affairs and are critical towards an addressee.
        
        \item \textbf{Irony:} tweets that contradict the state of affairs but are not obviously critical towards an addressee.
        
        \item \textbf{Satire:} tweets that appear to support an addressee, but contain underlying disagreement and mocking.
        
        \item \textbf{Understatement:} tweets that undermine the importance of the state of affairs they refer to.
        
        \item \textbf{Overstatement:} tweets that describe the state of affairs in obviously exaggerated terms.
        
        \item \textbf{Rhetorical question:} tweets that include a question whose invited inference (implicature) is obviously contradicting the state of affairs.
    \end{enumerate}

%%%%%%% BERT classification Architecture %%%%%
\subsection{BERT Classification Architecture} 
\label{BERT-arch}

\begin{figure}[h]
\centering
\includegraphics[width=\linewidth]{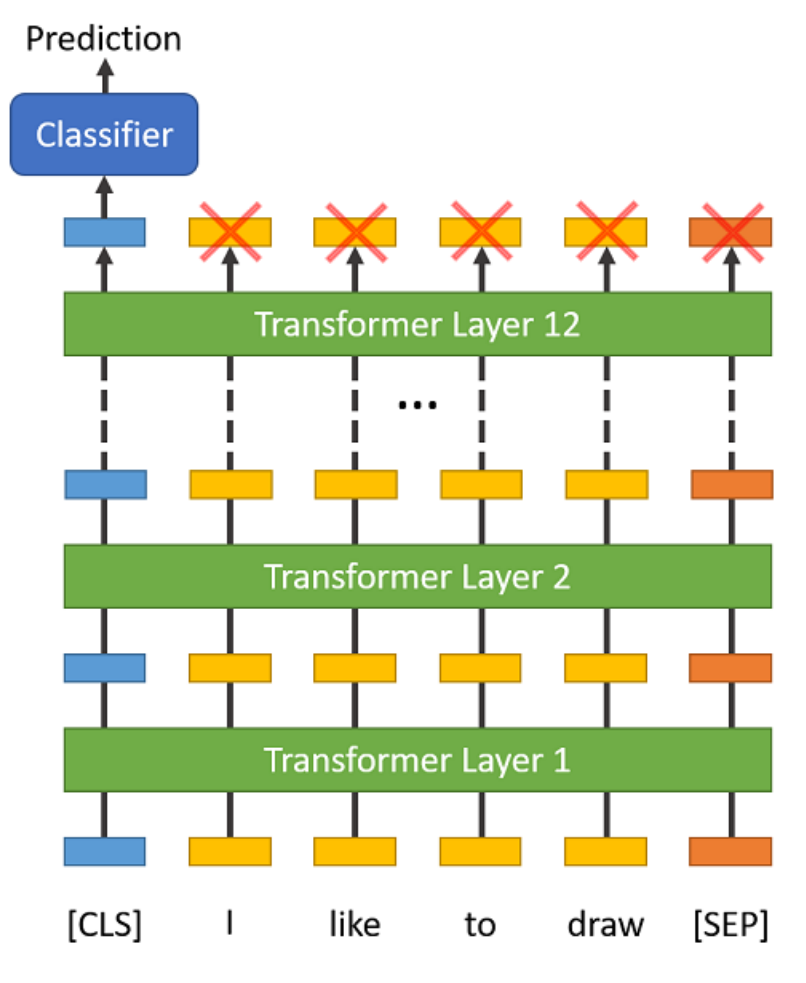}
\caption{BERT classification architecture}
\label{fig:BERT_classification_arch}
\end{figure}

Figure \ref{fig:BERT_classification_arch} shows the BERT-base classification architecture\footnote{\href{https://mccormickml.com/2019/07/22/BERT-fine-tuning/}{BERT Fine-Tuning Tutorial with PyTorch} by Chris McCormick and Nick Ryan}. 
From the output of the final (12th) transformer, only the first embedding (corresponding to the [CLS] token) is used by a classifier.

%%%%%%%%%%%%%%%%%%%%%%%%%%%%%%%%%%%

%%%%%%% ELECTRA %%%%%
\subsection{ELECTRA:- Replaced Token Detection}
\label{ELECTRA-arch}
ELECTRA (\textbf{E}fficiently \textbf{Le}arning an \textbf{E}ncoder that \textbf{C}lassifies \textbf{T}oken \textbf{R}eplacement \textbf{A}ccurately) \cite{clark2020electra} replaces the MLM of BERT with Replaced Token Detection (RTD), which looks to be more efficient and produces better results. In BERT, the input is replaced by some tokens with [MASK] and then a model is trained to reconstruct the original tokens.
    
In ELECTRA, instead of masking the input, the approach corrupts it by replacing some input tokens with plausible alternatives sampled from a small generator network. Then, instead of training a model that predicts the original tokens, a discriminative model is trained that predicts whether each token in the corrupted input was replaced by a generator sample or not.

\begin{figure*}[]
\centering
\includegraphics[width=0.8\textwidth]{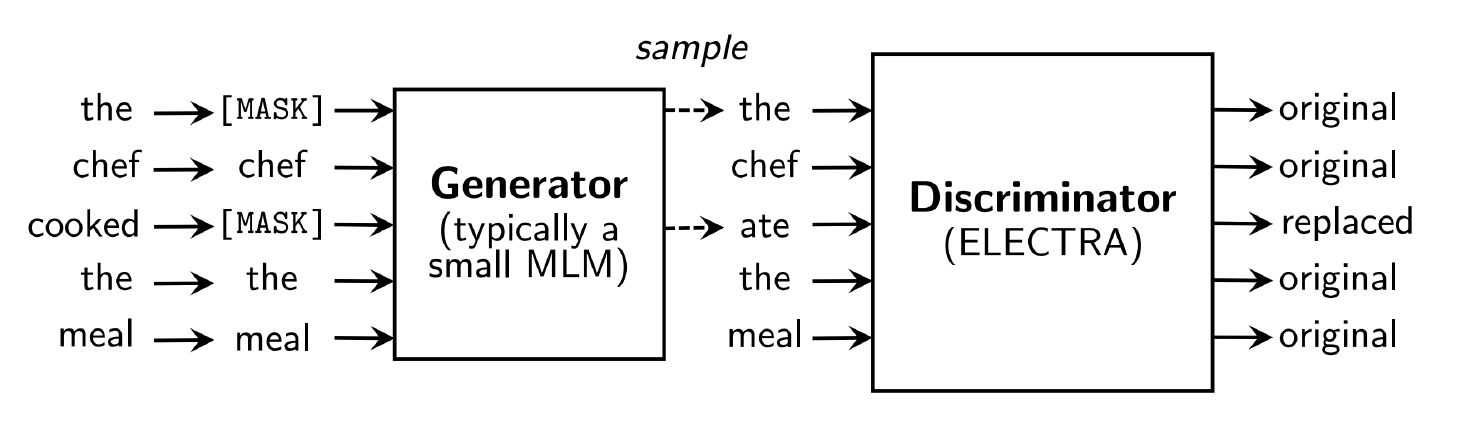}
\caption{An overview of replaced token detection}
\label{fig:ELECTRA_arch}
\end{figure*}

This approach trains two neural networks, a generator and a discriminator. Each one primarily consists of an encoder (e.g., a transformer network) that maps a sequence of input tokens into a sequence of contextualized vector representations. The discriminator then predicts whether it’s fake by analyzing its data distribution. An overview of replaced token detection is shown in Figure \ref{fig:ELECTRA_arch}.

%%%%%%%%%%%%%%%%%%%%%%%%%%%%%%%%%%%

\subsection{Effect of Learning Rates and Epochs in Subtask A (English)}
\label{Learning_Rates_Epochs}
In Subtask A (English), we experimented the BERT model on the B4 dataset with different learning rates and epochs as shown in Table \ref{tab:TaskA-En-Learning-Rates} and Table \ref{tab:TaskA-En-Epochs} respectively. As shown in the tables the best learning rate was 4e-6 and the best number of epochs was 5.

\begin{table}[h]
\centering
\begin{tabular}{|c|c|c|}
\hline
\rowcolor[HTML]{739CB9} 
{\color[HTML]{FFFFFF} \textbf{Val}} & {\color[HTML]{FFFFFF} \textbf{Test}} & {\color[HTML]{FFFFFF} \textbf{Learning Rate}} \\ \hline
\rowcolor[HTML]{FFFFFF} 
0.4002 & 0.25   & 1 e - 5 \\ \hline
\rowcolor[HTML]{FFFFFF} 
0.5871 & 0.4558 & 9 e - 6 \\ \hline
\rowcolor[HTML]{FFFFFF} 
0.5913 & 0.4639 & 8 e - 6 \\ \hline
\rowcolor[HTML]{FFFFFF} 
0.4002 & 0.25   & 7 e - 6 \\ \hline
\rowcolor[HTML]{FFFFFF} 
0.4002 & 0.25   & 6 e - 6 \\ \hline
\rowcolor[HTML]{FFFFFF} 
0.5952 & 0.4883 & 5 e - 6 \\ \hline
\cellcolor[HTML]{D5AB6D}0.6025      & \cellcolor[HTML]{6D9EEB}0.4769       & \cellcolor[HTML]{FFFFFF}4 e - 6              \\ \hline
\rowcolor[HTML]{FFFFFF} 
0.5798 & 0.475  & 3 e - 6 \\ \hline
\rowcolor[HTML]{FFFFFF} 
0.5644 & 0.4724 & 2 e - 6 \\ \hline
\rowcolor[HTML]{FFFFFF} 
0.6012 & 0.5017 & 1 e - 6 \\ \hline
\rowcolor[HTML]{FFFFFF} 
0.4002 & 0.25   & 9 e - 7 \\ \hline
\rowcolor[HTML]{FFFFFF} 
0.4002 & 0.25   & 8 e - 7 \\ \hline
\end{tabular}
\caption{The effect of learning rates on performance of the BERT model on the B4 dataset with weighted loss function}
\label{tab:TaskA-En-Learning-Rates}
\end{table}

\begin{table}[h]
\centering
\begin{tabular}{|c|c|c|}
\hline
\rowcolor[HTML]{739CB9} 
{\color[HTML]{FFFFFF} \textbf{Val}} & {\color[HTML]{FFFFFF} \textbf{Test}} & {\color[HTML]{FFFFFF} \textbf{Epochs}} \\ \hline
\rowcolor[HTML]{FFFFFF} 
0.5282 & 0.5304 & 1  \\ \hline
\rowcolor[HTML]{FFFFFF} 
0.5877 & 0.5092 & 3  \\ \hline
\cellcolor[HTML]{D5AB6D}0.6025 & \cellcolor[HTML]{6D9EEB}0.4769 & \cellcolor[HTML]{FFFFFF}5     \\ \hline
\rowcolor[HTML]{FFFFFF} 
0.6019 & 0.4647 & 7  \\ \hline
\rowcolor[HTML]{FFFFFF} 
0.5856 & 0.457  & 10 \\ \hline
\rowcolor[HTML]{FFFFFF} 
0.6017 & 0.5099 & 13 \\ \hline
\rowcolor[HTML]{FFFFFF} 
0.594  & 0.475  & 15 \\ \hline
\rowcolor[HTML]{FFFFFF} 
0.575  & 0.4943 & 17 \\ \hline
\rowcolor[HTML]{FFFFFF} 
0.5882 & 0.4675 & 20 \\ \hline
\rowcolor[HTML]{FFFFFF} 
0.5938 & 0.4633 & 23 \\ \hline
\rowcolor[HTML]{FFFFFF} 
0.575  & 0.4926 & 25 \\ \hline
\end{tabular}
\caption{The effect of epochs on performance of the BERT model on the B4 dataset with weighted loss function}
\label{tab:TaskA-En-Epochs}
\end{table}

\subsection{Performance of BERT+BiLSTM with and without Attention in Subtask A (English)}
\label{BERT-BiLSTM-appendix}

\textbf{BERT+BiLSTM:}
The best results we got using this architecture were on the B4 dataset with 4e-6 learning rate, 10 epochs, and 50 LSTM hidden state size. Table \ref{tab:TaskA-En-BERT-RNN} shows the results using same hyperparameters but with different hidden state sizes. 

\begin{table}[h]
\centering
\begin{tabular}{|c|c|c|}
\hline
\rowcolor[HTML]{739CB9} 
{\color[HTML]{FFFFFF} \textbf{\begin{tabular}[c]{@{}c@{}}Hidden \\ State Size\end{tabular}}} & {\color[HTML]{FFFFFF} \textbf{Val}} & {\color[HTML]{FFFFFF} \textbf{Test}} \\ \hline
50  & \cellcolor[HTML]{D5AB6D}0.6027 & 0.4741 \\ \hline
100 & 0.5882                         & 0.4765 \\ \hline
300 & 0.5813                         & 0.4627 \\ \hline
600 & 0.561                          & 0.4702 \\ \hline
900 & 0.5831                         & 0.4516 \\ \hline
\end{tabular}
\caption{Results of the BERT+BiLSTM model for SubTask A (English) on the B4 dataset}
\label{tab:TaskA-En-BERT-RNN}
\end{table}

\textbf{BERT+BiLSTM+Attention:}
The best results we got using this architecture were on the B3 dataset with 4e-6 learning rate, 5 epochs, and 600 LSTM hidden 
state size. Table \ref{tab:TaskA-En-BERT-RNN-ATT} shows the results using same hyperparameters but with different hidden state sizes. 

\begin{table}[h]
\centering
\begin{tabular}{|c|c|c|}
\hline
\rowcolor[HTML]{739CB9} 
{\color[HTML]{FFFFFF} \textbf{\begin{tabular}[c]{@{}c@{}}Hidden State \\ Size\end{tabular}}} & {\color[HTML]{FFFFFF} \textbf{Val}} & {\color[HTML]{FFFFFF} \textbf{Test}} \\ \hline
50  & 0.5726                         & 0.4685 \\ \hline
100 & 0.5978                         & 0.468  \\ \hline
300 & 0.5935                         & 0.4627 \\ \hline
600 & \cellcolor[HTML]{D5AB6D}0.6087 & 0.4625 \\ \hline
900 & 0.5777                         & 0.4618 \\ \hline
\end{tabular}
\caption{Results of the BERT+BiLSTM+Attention model for SubTask A (English) on B3 dataset}
\label{tab:TaskA-En-BERT-RNN-ATT}
\end{table}

\subsection{More Information about the Datasets:}
\label{More_Information}
\begin{figure}[h]
\centering
\includegraphics[width=\columnwidth]{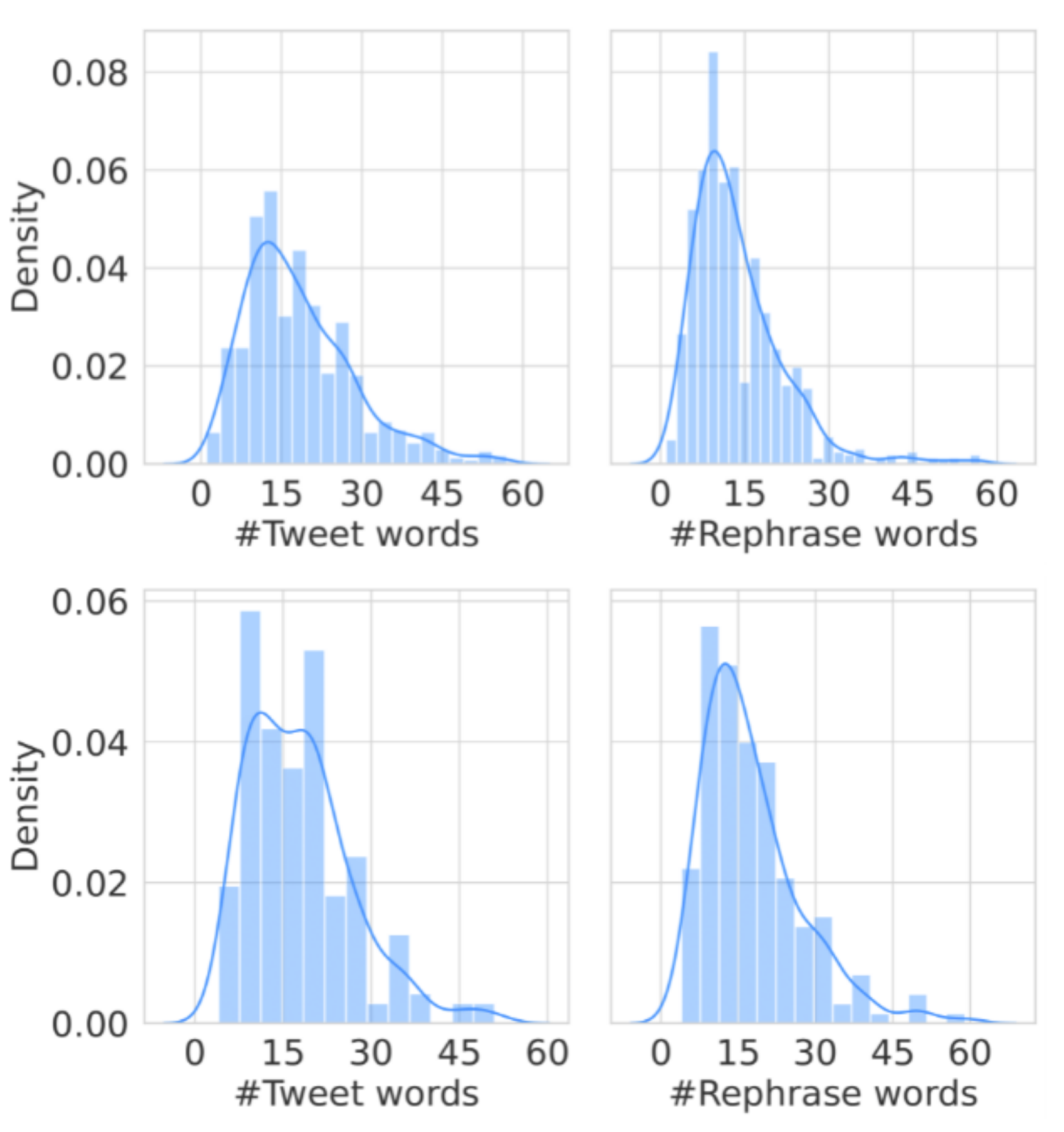}
\caption{Density of the number of words in the original English training (top) and testing (bottom) datasets.}
\label{fig:OriginalEnglishNumberOFWords}
\end{figure}

\begin{figure}[h]
\centering
\includegraphics[width=\columnwidth]{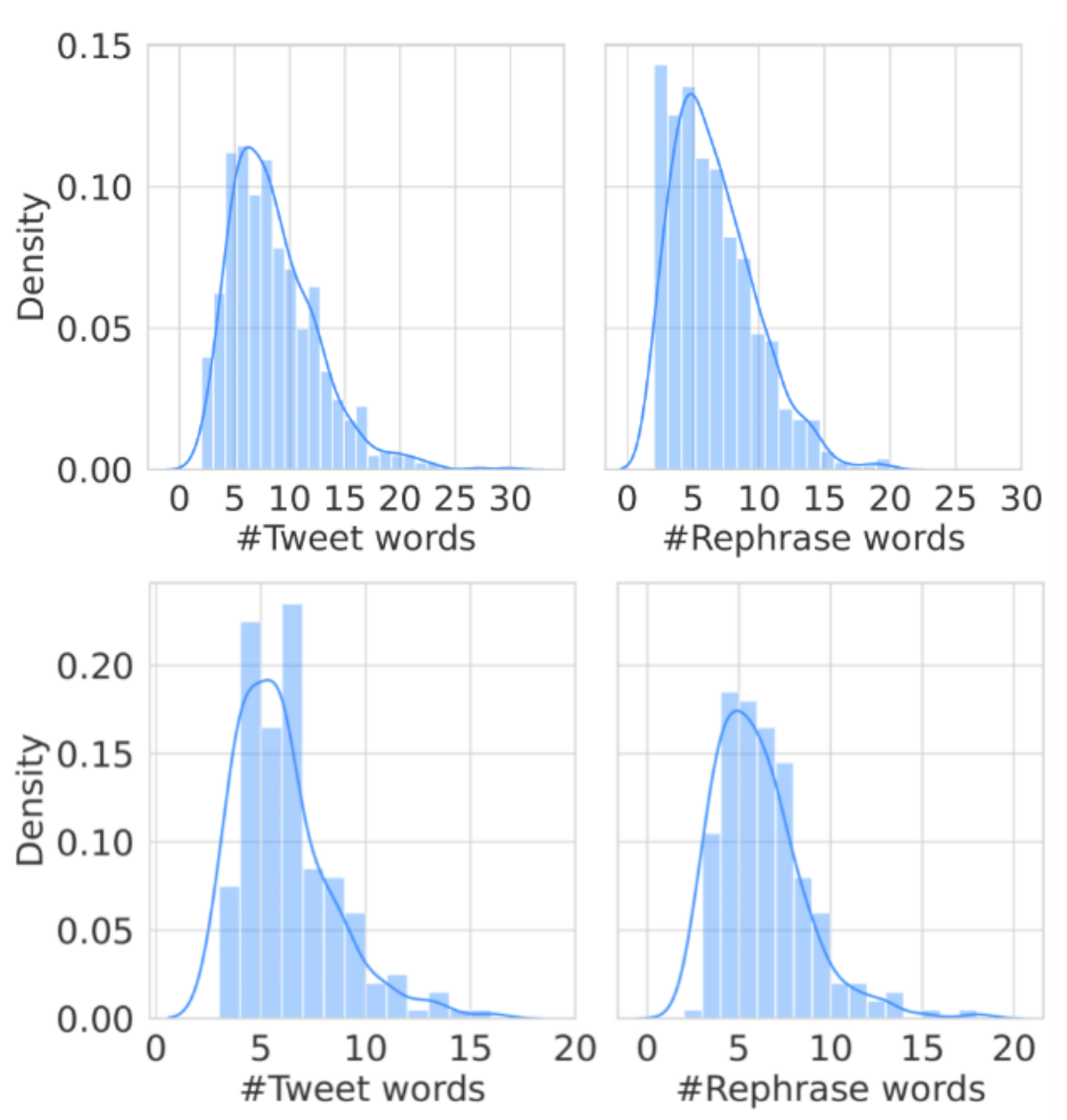}
\caption{Density of the number of words in the original Arabic training (top) and testing (bottom) datasets.}
\label{fig:OriginalArabicNumberOFWords}
\end{figure}

Density of the number of words in tweets and their rephrases in the original datasets is shown in Figure \ref{fig:OriginalEnglishNumberOFWords} for English and in Figure \ref{fig:OriginalArabicNumberOFWords} for Arabic.

\begin{figure}[h]
\centering
\includegraphics[width=\columnwidth]{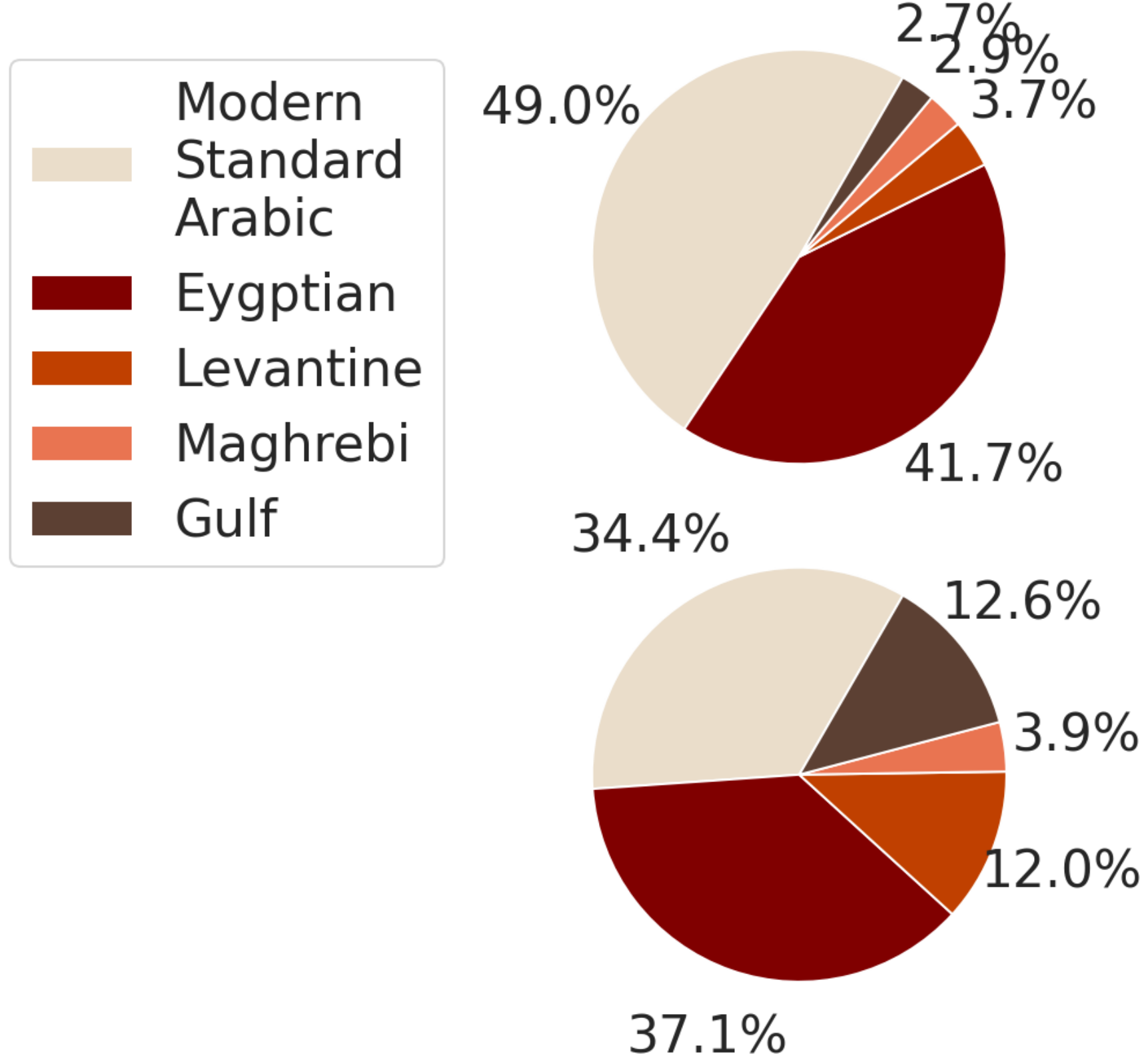}
\caption{Percentage of dialects of tweets in the original Arabic training (top) and testing (bottom) datasets.}
\label{fig:OriginalArabicDialects}
\end{figure}

In addition to this, information about dialects for Arabic subtasks is presented in Figure \ref{fig:OriginalArabicDialects} and Table \ref{tab:Dialects}.

\begin{table}[h]
\begin{tabular}{|c|c|c|c|c|}
\hline
\rowcolor[HTML]{739CB9} 
\cellcolor[HTML]{739CB9}{\color[HTML]{FFFFFF} } &
  \cellcolor[HTML]{739CB9}{\color[HTML]{FFFFFF} } &
  \cellcolor[HTML]{739CB9}{\color[HTML]{FFFFFF} } &
  {\color[HTML]{FFFFFF} \textbf{Eygptian}} &
  {\color[HTML]{FFFFFF} \textbf{MSA}} \\ \cline{4-5} 
\rowcolor[HTML]{739CB9} 
\cellcolor[HTML]{739CB9}{\color[HTML]{FFFFFF} } &
  \cellcolor[HTML]{739CB9}{\color[HTML]{FFFFFF} } &
  \cellcolor[HTML]{739CB9}{\color[HTML]{FFFFFF} } &
  {\color[HTML]{FFFFFF} \textbf{Levantine}} &
  {\color[HTML]{FFFFFF} \textbf{Gulf}} \\ \cline{4-5} 
\rowcolor[HTML]{739CB9} 
\multirow{-3}{*}{\cellcolor[HTML]{739CB9}{\color[HTML]{FFFFFF} \textbf{Dataset}}} &
  \multirow{-3}{*}{\cellcolor[HTML]{739CB9}{\color[HTML]{FFFFFF} \textbf{Split}}} &
  \multirow{-3}{*}{\cellcolor[HTML]{739CB9}{\color[HTML]{FFFFFF} \textbf{Total}}} &
  {\color[HTML]{FFFFFF} \textbf{Maghrebi}} &
  {\color[HTML]{FFFFFF} \textbf{}} \\ \hline
 &
  \cellcolor[HTML]{FFFFFF} &
   &
  41.7 &
  49 \\ \cline{4-5} 
 &
  \cellcolor[HTML]{FFFFFF} &
   &
  3.7 &
  2.7 \\ \cline{4-5} 
 &
  \multirow{-3}{*}{\cellcolor[HTML]{FFFFFF}Train} &
  \multirow{-3}{*}{3102} &
  2.9 &
   \\ \cline{2-5} 
 &
  \cellcolor[HTML]{FFFFFF} &
   &
  37.1 &
  34.4 \\ \cline{4-5} 
 &
  \cellcolor[HTML]{FFFFFF} &
   &
  12 &
  12.6 \\ \cline{4-5} 
\multirow{-6}{*}{Original} &
  \multirow{-3}{*}{\cellcolor[HTML]{FFFFFF}Test} &
  \multirow{-3}{*}{1400} &
  3.9 &
   \\ \hline
 &
  \cellcolor[HTML]{FFFFFF} &
   &
  21.2 &
  68.2 \\ \cline{4-5} 
 &
  \cellcolor[HTML]{FFFFFF} &
   &
  5 &
  5.1 \\ \cline{4-5} 
 &
  \multirow{-3}{*}{\cellcolor[HTML]{FFFFFF}Train} &
  \multirow{-3}{*}{12548} &
  0.3 &
   \\ \cline{2-5} 
 &
  \cellcolor[HTML]{FFFFFF} &
   &
  10.2 &
  77.4 \\ \cline{4-5} 
 &
  \cellcolor[HTML]{FFFFFF} &
   &
  1.6 &
  10.7 \\ \cline{4-5} 
\multirow{-6}{*}{\begin{tabular}[c]{@{}c@{}}ArSarcasm\\ -v2\end{tabular}} &
  \multirow{-3}{*}{\cellcolor[HTML]{FFFFFF}Test} &
  \multirow{-3}{*}{3000} &
  0.1 &
   \\ \hline
\end{tabular}
\caption{Distribution of tweets over dialects in Arabic Datasets}
\label{tab:Dialects}
\end{table}

\begin{figure}[h]
\centering
\includegraphics[width=\columnwidth]{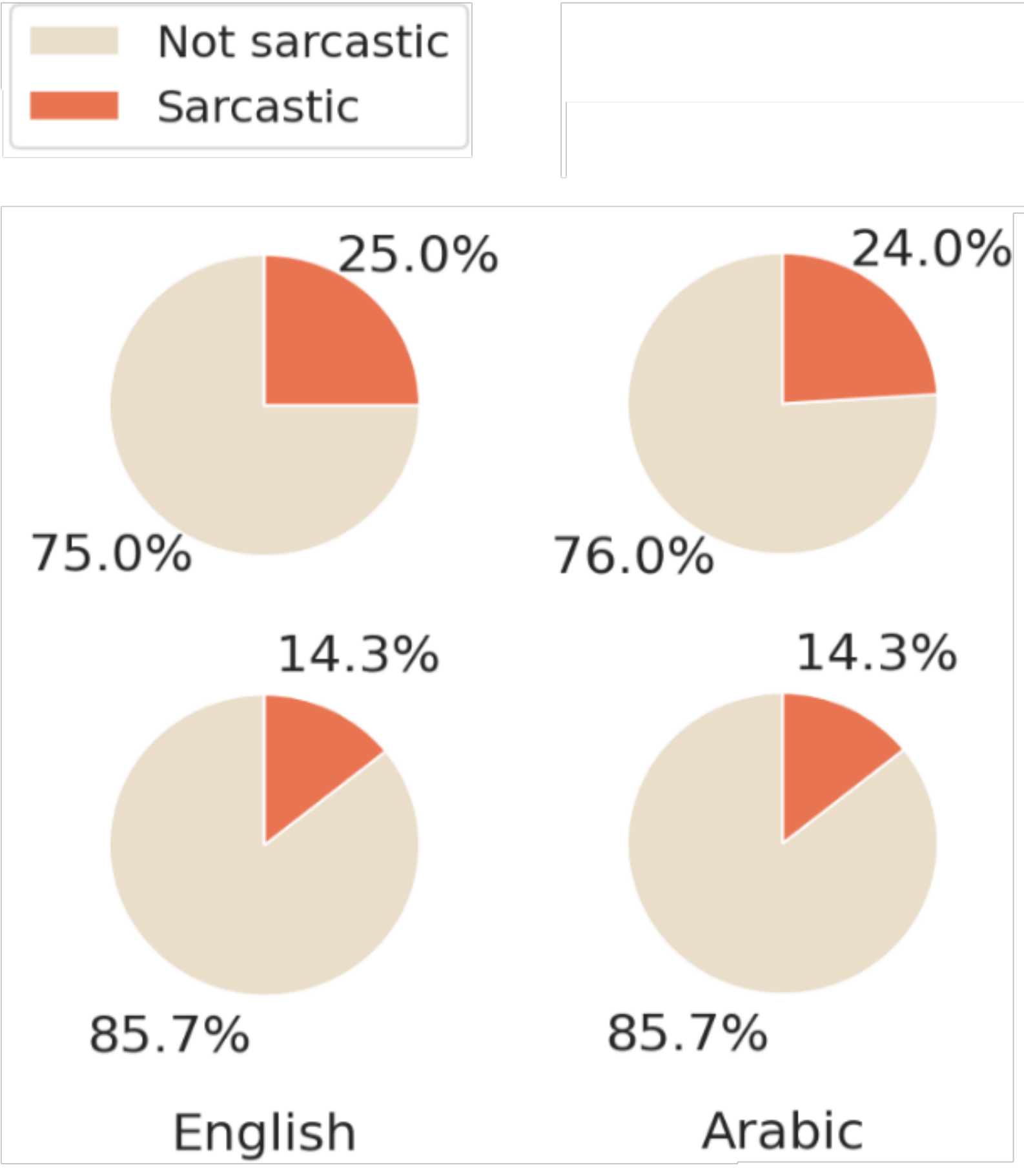}
\caption{Percentage of sarcastic and non-sarcastic tweets in the original English and Arabic training (top) and testing (bottom) datasets.}
\label{fig:OriginalEnglishArabicTaskAC}
\end{figure}

Information about the distribution of sarcastic and non-sarcastic tweets in the original datasets is presented in Figure \ref{fig:OriginalEnglishArabicTaskAC}.

\begin{figure}[h]
\centering
\includegraphics[width=\columnwidth]{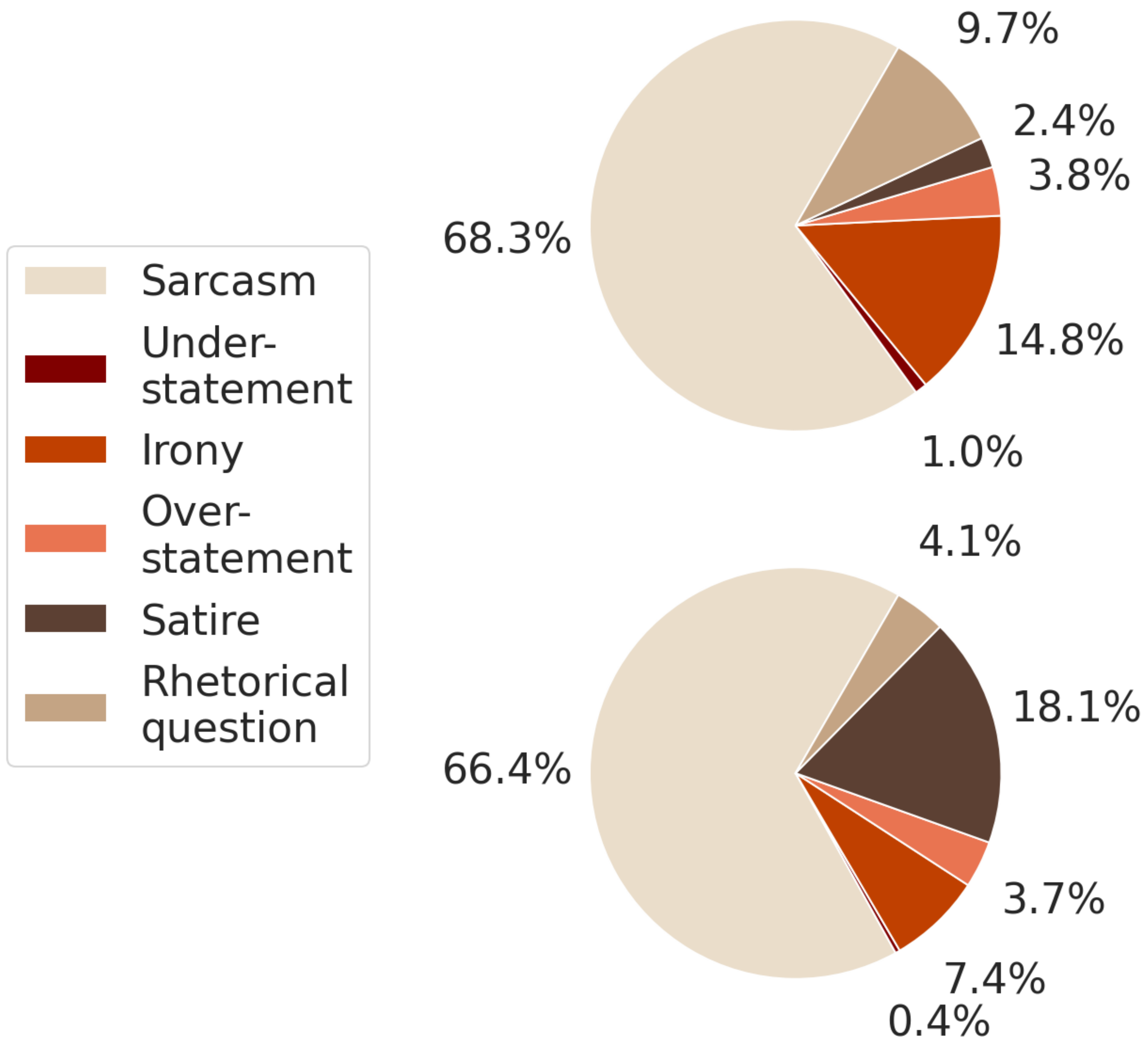}
\caption{Percentage of tweets under each sarcastic label in the original English training (top) and testing (bottom) datasets.}
\label{fig:OriginalEnglishTaskB}
\end{figure}

Information about sarcastic labels, for subTask B, in the original datasets is shown in Figure \ref{fig:OriginalEnglishTaskB}.

\end{document}